\pdfoutput=1
\PassOptionsToPackage{table,xcdraw}{xcolor}
\documentclass[sigconf]{acmart}

\usepackage[table,xcdraw]{xcolor}

\usepackage{pgfplots}
\usepackage{algorithmic}
\usepackage{algorithm}
\usepackage{amsmath}

\usepackage{amsfonts}
\usepackage{balance}
 \usepackage{booktabs}  
\usepackage{epsfig}
\usepackage{multirow}
\usepackage{xspace}
\usepackage{url}
\usepackage{color}
\usepackage{balance}
\usepackage{eqparbox}
\usepackage{mathrsfs}
\usepackage{caption}
\usepackage{subcaption}
\usepackage{lipsum}
\usepackage{enumitem}
\usepackage{float}
\usepackage{array, multirow}
\usepackage{balance}
\usepackage{hyperref}
\usepackage{mathtools}
\newcommand{\defeq}{\triangleq}
\usepackage{graphicx} \usepackage{bmpsize}
\usepackage{tabularx}
\usepackage{makecell}

\usepackage{soul}

\newcommand{\feicomment}[1]{{\color{orange} }}
\newcolumntype{L}[1]{>{\raggedright\let\newline\\\arraybackslash\hspace{0pt}}m{#1}}
\newcolumntype{C}[1]{>{\centering\let\newline\\\arraybackslash\hspace{0pt}}m{#1}}
\newcolumntype{R}[1]{>{\raggedleft\let\newline\\\arraybackslash\hspace{0pt}}m{#1}}

\newtheorem{Theorem}{Theorem}
\newtheorem{Definition}{Definition}

\newcommand{\eat}[1]{}

\definecolor{L_color}{rgb}{0.8,0.2,0.2}
\definecolor{Y_color}{rgb}{1.0, 0.0, 0.5}

\newcommand{\nop}[1]{}
\newcolumntype{Z}{ >{\centering\arraybackslash}X }

\newcommand{\yifan}[1]{}
\newcommand{\yifann}[2]{}

\newcommand{\jason}[1]{}
\newcommand{\madelon}{\texttt{madelon}}
\newcommand{\ana}{\texttt{a9a}}

\begin{document}
\title{
MetaCon: Unified Predictive Segments System with Trillion Concept Meta-Learning 
\\
}

\author{
Keqian Li$^{1}$, Yifan Hu$^{1}$, Logan Palanisamy$^{1}$,
Lisa Jones$^{1}$, 
Akshay Gupta$^{1}$, 
Jason Grigsby$^{1}$, 
Ili Selinger$^{2}$, 
Matt Gillingham$^{3}$, 
Fei Tan$^{4}$\\
Yahoo Research$^{1}$, JP Morgan \& Co $^{2}$, U.S. Bancorp $^{3}$, 
Xiaohongshu$^{4}$
}

\newcommand{\shortauthor}{K. Li et al.}
\renewcommand{\shorttitle}{MetaCon}

\begin{abstract}

Accurate understanding of users 
in terms of predicative segments
play an essential role in the day to day operation
of modern internet enterprises.
Nevertheless, there are significant challenges that 
limit the quality of data, especially on long tail predictive tasks.
In this work, we present \textit{MetaCon}, our unified predicative segments system with scalable, trillion concepts meta learning 
that addresses these challenges.
It builds on top of a flat concept representation \cite{li2021hadoop} that summarizes entities' heterogeneous digital footprint, 
jointly considers the entire spectrum of predicative tasks as a single learning task, and 
leverages principled meta learning approach 
with efficient first order meta-optimization procedure under a provable performance guarantee
in order to solve the learning task.
Experiments on both proprietary production datasets and public structured learning tasks 
demonstrate that \textit{MetaCon} can lead to substantial improvements over state of the art recommendation and ranking approaches.

\end{abstract}
\maketitle

\begin{figure}[]
\centering
\includegraphics[  width=.5\textwidth]{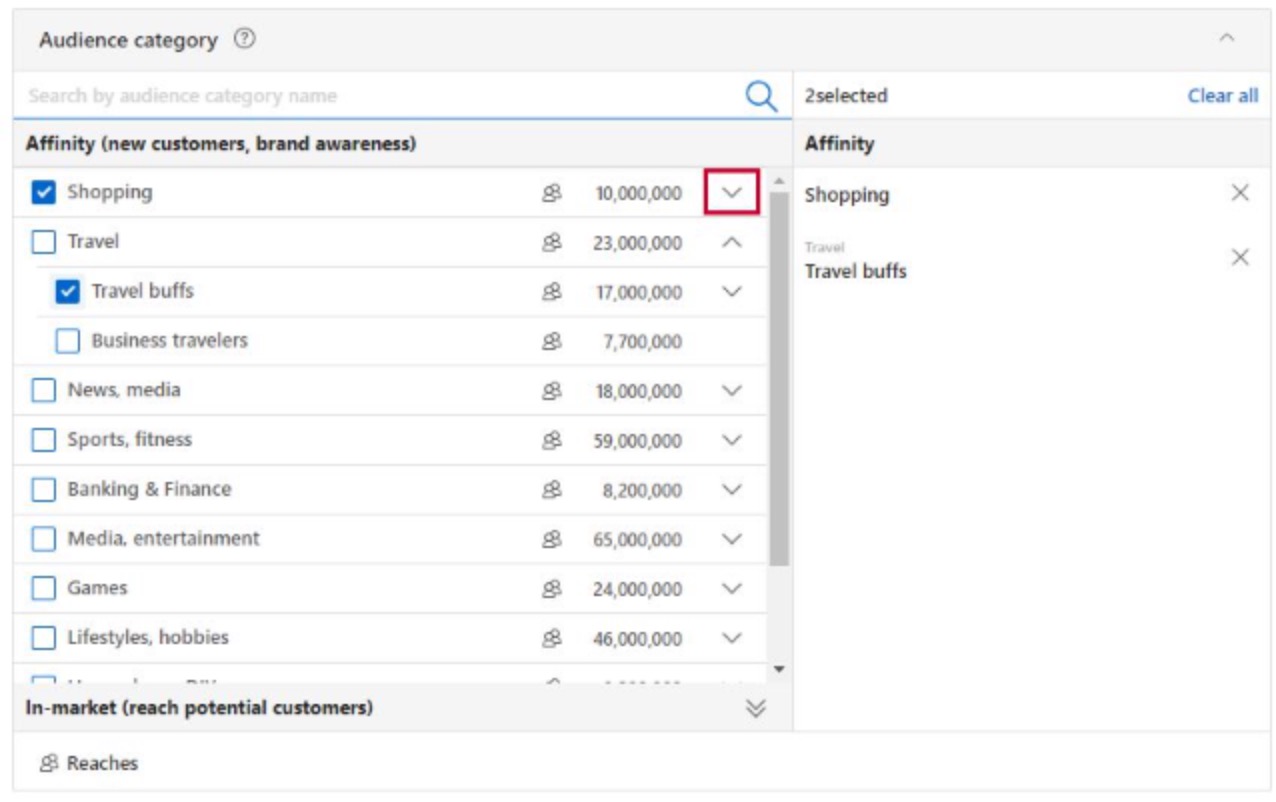}
\caption{Illustration of MetaCon use cases. Shown in the figure is a typical interface for advertisers to specify target audience segments. In this case "Shopping" and "Travel buffs" segments are selected.}

\label{fig-use case}

\end{figure}

\section{Introduction}

Ever since the introduction of large scale information service such as AOL, Yahoo, and Google,
along with scalable distributed storage and computation engines such as Hadoop, 
 accurate user understanding and content serving has been a crucial technology, 
where a slight variation of the model performance may result in significant downstream impact in user satisfaction and revenue \footnote{\url{https://news.yahoo.com/why-facebook-parent-metas-stock-is-getting-crushed-141137947.html}} \cite{balusamy2021driving}.
A typical paradigm adopted by most major players in the industry is 
AI based market segmentation, where a  dedicated predictive system is employed to predict segments: groups of users with shared characteristics.
As illustrated in \autoref{fig-use case}, 
a user's interest in particular categories such as sports, news, or travel can be predicted. Those predictions can be leveraged for the company to create more effective content
\cite{cahill1997target}
\footnote{\url{https://www.facebook.com/business/ads/ad-targeting}} \footnote{\url{https://support.google.com/google-ads/answer/2497941?hl=en}}.

Despite its critical nature, 
there exists several significant challenges for building the reliable predictive segments systems.
The first stems from \textit{irregularity of data},
as user's online activities are gathered from disparate domains and modality, with significant volumes 
both in terms of the user population and feature dimensionality.
The second challenge is about \textit{scarcity of signals},
where the
majority of features are expected to be absent due to the implicit feedback nature of the collected online activity. 
In a online world where the ``right to be forgotten'' \cite{voigt2017eu} is a prerequisite for continued and active service of a product,
privacy oriented measures for explicit user consent, desktop and browser cookie restrictions such as \textit{Chromageddon}  \cite{chromium},
and mobile analytics restrictions such as \textit{App Tracking Transparency} \footnote{\url{https://developer.apple.com/documentation/apptrackingtransparency}} and \textit{Intelligent Tracking Prevention} 
\cite{iqbal2021towards}, 
has made  the efficient utilization of data and knowledge the topmost priorities among major industry players.

Last but not the least,
the scarcity of signals also stems from the \textit{long tail distribution} of the large number of predicative segments tasks that are not "born equal". 
Due to the large quantity of possible segments \cite{cahill1997target}, there are many niche segments that may not expand to a large population and therefore lacks data and modeling resource but are nonetheless important for the corresponding users.

Current recommender systems \cite{zhao2019recommending} 
that rely on transfer learning and multi-objective optimization  
can not address the above challenges
because 
the tasks may not be well aligned as a video's like-comment-share statistics.
As is commonly seen in production, ETL pipelines for individual tasks
will go through completely different distributed database query,
with possibly no instances or features in common.
The importance of a reliable, accurate predictive segments system in the presence of the above scenario is extremely valuable.

To this end, we present \textit{MetaCon}, 
  our scalable unified predicative system that 
  builds on distributed concept representation of heterogeneous knowledge, and 
  jointly models the entire set of tasks as a unified predictive segments task,
  and leverages principled meta learning to efficiently share and accumulate knowledge across the component tasks,
  while at the same time retaining the flexibility for task-specific domain, data-set and optimization methods.

Our contribution can be summarized as follows
\begin{itemize}

     \item   We study the novel problem of jointly predicting user segments under unified modeling.
     \item   We present \textit{MetaCon} as an end to end solution that scales to trillions of concepts using an efficient and novel semi-synchronous meta learning algorithm with provable advantage over alternatives such as single task learning, vanilla multi task learning as well as competitive higher order gradient and approximate gradient family of meta learning algorithms.
     \item   We conduct extensive evaluations as well as ablation studies of \textit{MetaCon} over a large number of predicative segments tasks, and demonstrate that it
significantly outperforms the state of the art recommendation and ranking approaches, and that the meta learning component effectively learns crucial information to improve the overall performance on critical tasks.
     \item  We conduct general single task meta learning extension and evaluations on structured public datasets to further demonstrate the generalization of our approach.

\end{itemize}

\begin{figure*}[]
\centering
\includegraphics[width=\textwidth]{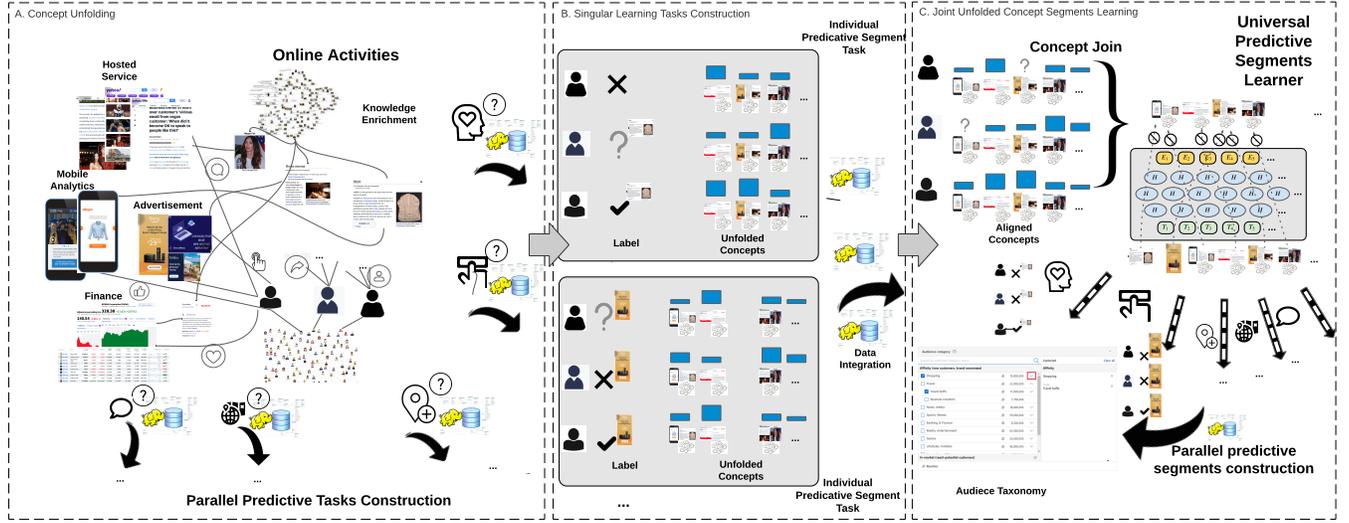}
\caption{ Overview of the \textit{MetaCone} system.
Online activities are integrated from heterogeneous sources 
from which tasks for various predictive segments are constructed. 
By cleaning and normalizing data for individual tasks and producing a aligned representation for tasks as a whole, 
a universal modeling system can be trained and deployed in a joint manner.}
\label{fig-illustration}
\end{figure*}

\vspace{-10pt}
\section{Related Work}
In this section,
we discuss key related works and their relations to \textit{MetaCon} for the three following categories: 
personalization and recommendation system, 
concept mining and concept representation, 
and meta learning.

\subsection{Personalization and Recommendation System}
In typical online advertising and recommendation systems
 \cite{mcmahan2013ad},
 \cite{liu2017related,he2014practical, zhai2017visual, freno2017practical, covington2016deep, }, 
content is served to an end user in a personalized fashion with the goal of maximizing downstream objectives, including engagement, 
with multiple previous works focusing on the problem of multi objective recommendation, 
\cite{lu2018like,agarwal2011localized,
wang2016multi,}
and in particular recent work  \cite{zhao2019recommending,ma2018entire, tang2020progressive} on  multi-task learning with deep neural networks. Our paper  generalizes this idea to allow for arbitrary unaligned tasks that do not share sample spaces to benefit from each other.

\subsection{Concept Learning}
The research on concept learning \cite{fischbein1996psychological, li2019mining} focuses on obtaining and exploiting 
semantically meaningful information from noisy structured data
under set expansion \cite{wang2015concept}  or hierarchical clustering  \cite{li2014social,li2017discovering}  paradigm, 
or
from unstructured text data \cite{zha2018fts, li2018poqaa,zha2019mining,li2019hiercon} by leveraging distributed semantics \cite{li2018unsupervised, li2018concept}.
We build on top of previous work \cite{li2021hadoop} and leverage the shared representation between concepts and tasks to improve downstream applications.

\subsection{Meta Learning}
The research of meta learning, also known as "learning to learn", focuses on 
training models that gain experience and improve performance over multiple learning episodes/tasks 
\cite{thrun1998learning},
which has seen wide adoption in research areas such as 
reinforcement learning \cite{kirsch2019improving, schweighofer2003meta}.
This can be applied in various aspects of learning problems under a bi-level optimization framework \cite{franceschi2018bilevel}, 
including the data set generation \cite{cubuk2018autoaugment},
learning objective \cite{zhou2020online, bechtle2021meta, kirsch2019improving},
model architecture \cite{liu2018darts}, 
initialization parameters \cite{finn2017model}, 
and the learning rules \cite{schweighofer2003meta,bengio1990learning, }.
Our work studies the meta learning in the context of large scale segment prediction tasks, and further investigates meta learning's application to single task learning with novel auxiliary task family construction.

\newcommand{\task}{\mathcal{T}}
\newcommand{\loss}{\mathcal{L}}
\newcommand{\inp}{\mathbf{x}}
\newcommand{\learner}{f}
\newcommand{\lossi}{\loss_{\task_i}}

\section{Problem Overview}
The system of Unified Predictive Segments can be illustrated in \autoref{fig-illustration},
where different items of interest from disparate domains such as Hosted Content, Mobile, Advertisement and Finance
are collected as raw events, 
further enriched with mined entity level association with knowledge base,
 and augmented with users, events and existing segments.
As a result, the input to our problem is large collection of datasets where 
each individual component is of arbitrary schema with arbitrary relation associations among the objects.
We start by following previous distributed AutoML approach for data integration and transform each task for predicting a specific segment into an equivalent \textit{unfolded concept learning problem} \cite{li2021automl} in parallel.

Formally, we assume there are a set of $K$ segment prediction tasks, the $i$-th task is represented by $\mathcal{T}_i$, with $\mathcal{T}_i \in \mathbb{T}$.
For each task $\mathcal{T}_i$, let the sample space $\mathcal{S}_i$ be the set of users that we predict the segments for
and $\mathcal{Y}_i$ be the set of possible labels.
 By leveraging \textit{concept unfolding} for cleaning and pre-processing,
different types of relations 
between entities and literals 
based on a meta-path \cite{yang2020heterogeneous} or SPARQL query
\jason{SparkSQL?}
are serialized as an atomic concept
\cite{li2021automl}.
Consequentially, 
there will be a concept vector $(\vec{c_s})_i$ 
for each user $s$ mapping each element in the list of concept vocabulary $\mathcal{C}_i$ into a real number.
For clarity, we start by formally stating the singular unfolded concept learning task as follows.

\begin{Definition} [Singular Unfolded Concept Learning] \label{def concept}
Assuming the label function of interest
$\mathbf{y}_i:\mathcal{S}_i \to \mathcal{Y}_i$ mapping each user to a label in $\mathcal{Y}_i$,
the task is to learn a model $\mathbf{h}_i \in \mathcal{H}_{i} \subseteq \mathbb{R}^{\mathcal{C}_i} \to \mathcal{Y}_i$,
that minimize the expected risk
according to a given criteria $\mathbf{L}_i$
under a probability measure of the entity $\mathbf{q}_i : \mathcal{S}_i \to [0,1]$:

\begin{equation*}
\begin{aligned}
& \underset{\mathbf{h}_{i}}{\text{minimize }}
R_i(h_{i} )
\defeq \mathbb{E}
_{\mathbf{q} _i}[\mathbf{L}_i(
h_i(
\vec{c_s})_i; \theta_i, \omega
), \mathbf{y}(s)
)]\\
&  = \int_{\mathcal{S}_i}  
\mathbf{L}_i(
h_i(
\vec{c_s})_i; \theta_i, \omega
), \mathbf{y}(s)
)
\mathbf{q} _i(s) d s\\
\end{aligned}
\end{equation*}
where $\theta_i \in \Theta_i$ denotes the task specific parameter, such as weights of neural network,
and $\omega \in \Omega$ denotes parameter for encoding the dependence on the assumptions about `how to learn', such as the choice of hyper-parameter of model architecture \cite{bayer2009evolving, zoph2016neural, franceschi2018bilevel}, initialization \cite{finn2017model} or optimizer for $\theta_i$ \cite{bengio1990learning, metz2018meta, schweighofer2003meta}.
\end{Definition}

As opposed to the conventional assumption that each minimization problem is solved from scratch and that $\omega$ is pre-specified, here we aim to improve the performance for individual minimization problem by \textit{learning} the parameter $\omega$ controlling the learning process for individual learning task.
To accomplish this, the ``meta knowledge'' $\omega$ will be learned to improve downstream task specific performance
by considering how the expected risk on individual tasks from each distribution $\mathbf{q}_i$ changes with respect to the meta knowledge. In effect, the training set of individual task becomes the test set for meta knowledge to adapt.

We start by formalizing this meta-learning problem setting from a general point of view.
Consider a distribution over tasks $\mathbf{p}: \mathbb{T} \to [0,1]$,
we assume a \emph{set} of $M$  source training (i.e. the support in meta learning literature \cite{bechtle2021meta}) and 
validation (query in meta learning literature \cite{bechtle2021meta}) data-sets 
available sampled from $\mathbb{T}$,
$\mathscr{D}_{source} \defeq \{(\mathcal{D}^{train~(i)}_{source}, \mathcal{D}^{val~(i)}_{source}\}_{i=1}^M$, each consisting of members  corresponding to i.i.d. samples drawn the distribution of instances $\mathbf{q}_i$ of task $\mathbf{T}_i$ for the meta-training stage, and the goal is to learn from $\mathscr{D}_{source}$ the meta-knowledge that minimize the expected risks of downstream tasks.

Specifically, we denote the set of $Q$ target tasks used in the meta-testing stage as $\mathscr{D}_{target} \defeq \{\mathcal{D}^{train~(i)}_{target}, \mathcal{D}^{test~(i)}_{target}\}_{i=1}^Q$,
each consisting of members corresponding to i.i.d. samples drawn from the instance distribution $\mathbf{q}_i$ from task $\mathbf{T}_i$,
we use the previous acquired meta-knowledge $\omega$ to learn and minimize the empirical risk of each dataset
in the hope of minimizing the loss on the hold out set $ \{ \mathcal{D}^{test~(i)}_{target}\}_{i=1}^Q$:
\vspace{-10pt}
\begin{align} \label{eq:metatest}
&
 \theta^{*}_i (\omega)
 \defeq \arg\min_{\theta_i} R^{\mathcal{D}^{train~(i)}_{target} }_i(h_{i}(\cdot;\theta_i, \omega) )  \nonumber\\
& = \sum_{s \in \mathcal{D}^{train~(i)}_{target} }
\mathbf{L}_i(h_i(\vec{c_s}; \theta_i, \omega), \mathbf{y}_i(s))
\end{align}

For predictive segment tasks, we are not necessarily constrained to keep the source and target sets separate.
Specifically, we collect the task instances for every available task as the source dataset, for  meta-training,
and the target set, for meta-test and deployment. Formally, we assume a sample of data
$\mathscr{D} \defeq \{\mathcal{D}^{train~(i)},$ $\mathcal{D}^{val~(i)}, \mathcal{D}^{test~(i)}\}_{i=1}^K$ 
weighted by task distribution $\mathbb{T}(i)$,
and task specific instance distribution $\mathbf{q}_i(s)$
for each of the $K$ segment prediction tasks. 
We will use all $K$ tasks for training, i.e. $\mathscr{D}_{source} \defeq \{\mathcal{D}^{train~(i)}, \mathcal{D}^{val~(i)}\}_{i=1}^K$ and 
also use the $K$ tasks for meta-test, i.e. $\mathscr{D}_{target} \defeq \{\mathcal{D}^{train~(i)}, \mathcal{D}^{test~(i)}\}_{i=1}^K$. 
The Universal Predictive Segments Learning problem 
can then be defined as joint learning over $K$ individual unfolded concept learning tasks

\begin{Definition} [Universal Predictive Segments Learning]
\label{def concept}
Assuming there are $K$ tasks, 
where each task $\mathcal{T}_i$, $1 \leq i \leq K$, is associated with possibly different
set of users $\mathcal{S}_i$,
label function $\mathbf{y}_i$,
concept vocabulary $\mathcal{C}_i$, and input dataset
$\mathscr{D}  \defeq \{ (\mathcal{D}^{train~(i)}, \mathcal{D}^{val~(i)},  \mathcal{D}^{test~(i)}\}_{i=1}^K$ for entire  $K$ segment prediction tasks, the optimization goal is

\begin{align}
& \underset{}{\text{minimize }}
R^{meta}_{\{\mathcal{H}_{i}\}^K_{i=1}, \Omega}
\defeq \nonumber\\
& = \sum^{K}_{i=1} R^{\mathcal{D}^{test~i} }_i(h_{i}(\cdot;\theta^{*~(i)}(\omega^*), \omega^*) ) \nonumber\\
 & = \sum^{K}_{i=1} \left(
 \sum_{s \in \mathcal{D}^{test~i} }
\mathbf{L}_i(h_i(\vec{c_s}; \theta^{*~(i)}(\omega^*), \omega^*), \mathbf{y}_i(s))\right)
 \label{eq:con-meta}\\
 \text{s.t. } &\omega^* = \arg\min_{\omega} \mathcal{L}^{meta}(
 \theta^{*~(i)}(\omega), \omega, \mathcal{D}_{source})
\end{align}

\noindent where the hypothesis space is $\{\mathcal{H}_{i}\}^K_{i=1}, \Omega$ for the task-specific, meta parameters,
$\theta^{*~(i)} (\omega)$, specifies the adaptation procedure as defined in \autoref{eq:metatest}, and
$\mathbf{L}^{meta}$ is a meta loss to be specified by the meta-training procedure, such as cross entropy in the case of few-shot classification \cite{finn2017model}.

\end{Definition}

The unified predictive segments abstraction inherits the scalability and efficiency of concept learning  while retains the representation power for wider ranges of problems such as graph learning and heterogeneous information networks \cite{yang2020heterogeneous} which can be included as a special case in the meta learning solution space. The following results can be shown by reduction from unfolded concept learning problem in \cite{li2021automl}.
\begin{Theorem}
The above Universal Predictive Segments Learning problem is no less difficult than
$K$ arbitrary combinations of
Learning In Heterogeneous Data Problem (Definition 1 in \cite{li2021automl}),
Learning In Relational Database (Definition 2 in \cite{li2021automl}),
Heterogeneous Graph Learning (Definition 3 in \cite{li2021automl}), and
First Order Logic Graph Learning (Definition 4 in \cite{li2021automl}).
In fact, there exists efficient linear time reduction from $K$ arbitrary combinations of
Learning In Heterogeneous Data Problem,
$K$ Learning In Relational Database,
$K$ Heterogeneous Graph Learning, and
$K$ First Order Logic Graph Learning.
\end{Theorem}

\yifan{Put out an appendix??}

    \begin{figure*}[]
    \centering
    \includegraphics[width=\textwidth]{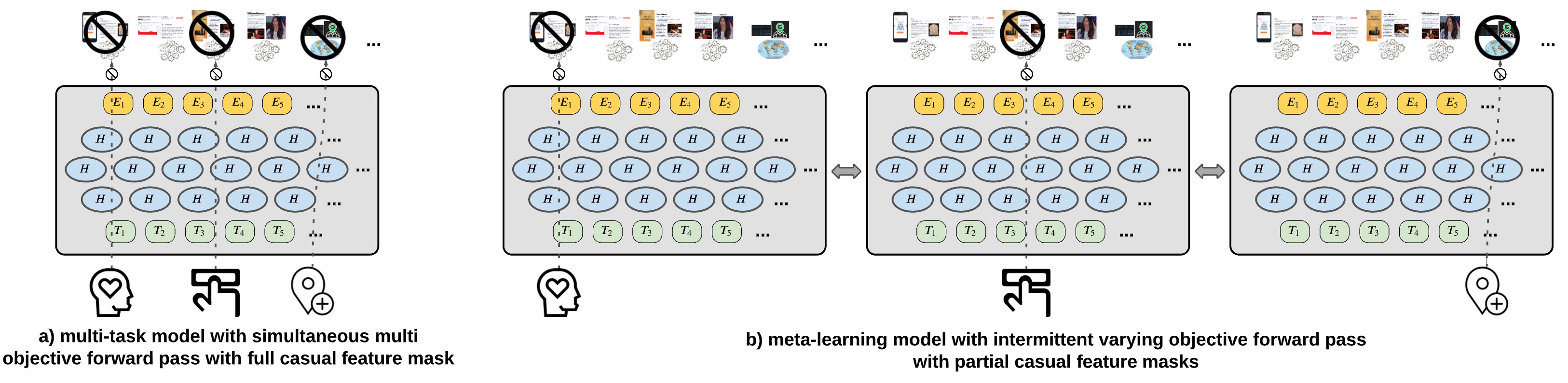}
    \caption{Comparison between \textit{MetaAug} and traditional multi-task learning paradigm.. 
    In multi-task learning (Sub-figure a), one single learner have to ignore the relevant information for all label at the same time when performing learning.
    In \textit{MetaAug}  (Sub-figure b), different learners are allowed to share a large common vocabulary and retain relevant information for other tasks for trying to learn from its own task}
    \label{fig-paradigm}
    \end{figure*}
    
    \begin{figure}[]
    \centering
    \includegraphics[width=.5\textwidth]{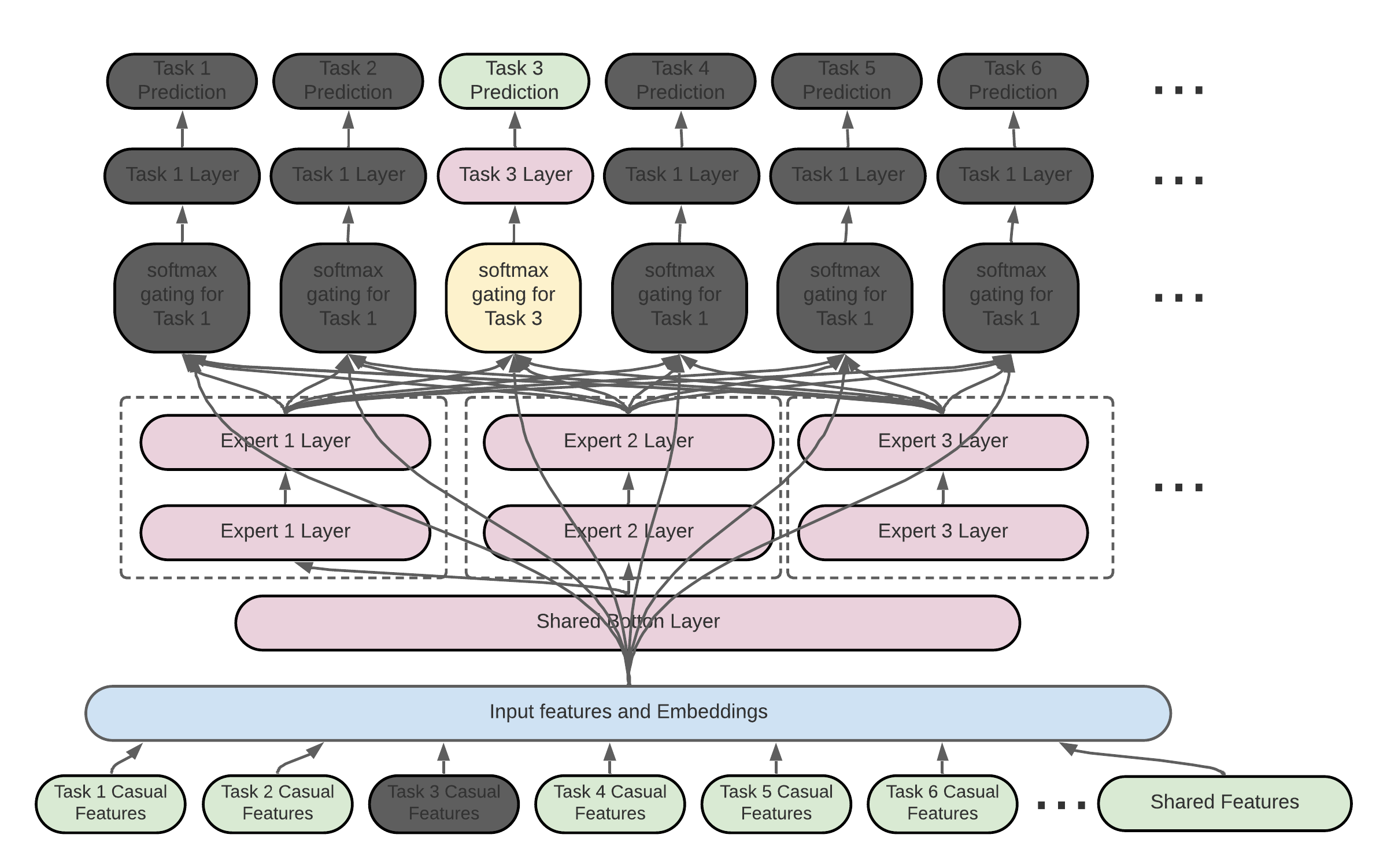}
    \caption{Illustration of meta optimization procedure on data instance from specific task. All features except those belong to causal mask for Task 3 are fed into shared experts and the output of the  head corresponding to Task 3 is activated for downstream prediction}
    \label{fig-network}
    \end{figure}
    \section{Choice of $\Omega$ and $\Theta_i$}
    A typical meta learning algorithm can be characterized by two major components, the representation of meta model and the optimization procedure for learning the model. 
    In this section we focus on the first part, specifically the meta parameter space $\Omega$ and the parameter space $\Theta_i$ for each task $\mathcal{T}_i$.
    
    The role of meta parameter $\omega$ is to impose an inductive bias on the  model landscape for each specific task. For deploying into critical scenarios such as predictive segments, 
    the following challenges arise for the learning system:
    \begin{enumerate}
        \item \textbf{Representation Power} 
        The choice of $\Omega$ should allow for large enough solution space of each single task $\Theta_i$ instead of limiting it to a specific function class to account for the complexities of tasks. 
        \item \textbf{Task Heterogeneity} 
        The choice of $\Omega$  should 
        \textit{simultaneously} allow 
        task specialized models $\Theta_i$ for different downstream tasks $\{\mathcal{T}\}_{i=1}^M$ with heterogeneous domain, modality, and different concept vocabulary $\{\mathcal{C}_i\}_{i=1}^M$.
        \item \textbf{First Order Influence} 
        The influence from meta parameter $\omega$ on the each of the task specific model parameters $\theta_i$ and resulting model $h_i(\cdot; \theta_i, \omega)$ should be analytically\jason{analyzed?} without second order\jason{second order what?} (single step MAML \cite{finn2017model}) to allow for efficient large scale optimization.
    \end{enumerate}
    We found that no previous learning approaches satisfy the above criteria. 
    Gradient-based meta learning that estimates the influence of meta parameter
    via approximations such as few steps of higher gradient descent with learned prior \cite{finn2017model} \cite{franceschi2018bilevel} are disqualified by (2) and (3). 
    Reinforcement learning and evolution algorithm based meta learning that relies on heuristics for estimating and
    optimization of the learner \cite{cubuk2018autoaugment, salimans2017evolution} is disqualified by (3).
    Black-box approaches that learn to ingest entire datasets to output final learners in a forward pass \cite{ravi2016optimization} and prototypical networks \cite{snell2017prototypical}
    are limited to simpler and specific architecture and are disqualified by (1). 

    Therefore, we present our \textit{MetaAug} architecture that achieves the above three properties. The intuition is to jointly align the tasks as well as their associated concepts, 
    learn transferable meta-representations that shared the same solution space with task specific learners, 
    while at the same time respecting domain and schema constraints for task heterogeneity.

    \noindent\textbf{Concept Alignment} In the first stage of \textit{concept alignment},
    we create a meta-task $\mathcal{T}^{meta}$ that combines the label spaces
    $\mathcal{Y}_i$ and resulted unfolded concept vocabulary $\mathcal{C}_i$ into a single solution space.
    Specifically, 
    we define 
    its label space $\mathcal{Y}^{meta} \defeq \Pi_{i=1}^M \mathcal{Y}$ as a join over every individual task,
    its concept vocabulary $\mathcal{C}^{meta}$, s.t.  
    $ \mathcal{C}_i\subseteq \mathcal{C}^{meta} $, 
    $\mathcal{Y}_i \subseteq \mathcal{C}^{meta} $, $\forall 1\leq i\leq M$ as a dictionary of all concepts and labels,
    and a meta level data-set $\mathcal{D}^{meta}$ such that every $\mathcal{D}^{train~(i)} \in \mathscr{D}$ is contained inside by
    $$ \mathcal{D}^{train~(i)} \leftarrow 
    \Pi_{
    (\{c_1, \ldots c_{|\mathcal{C}_i|} | c_i \in \mathcal{C}_i \} \bigcup \mathcal{Y}_i)
    } 
    (\sigma_{ (s \textrm{ in  } \mathcal{S}_i ) } 
    \mathcal{D}^{meta} )$$
    where the $\Pi(\cdot)$ and $\sigma(\cdot)$ only denote the \textit{project} and \textit{select} operators in relational algebra \cite{codd2002relational}, respectively.
    
    \noindent\textbf{Concept Augmentation With Casual Mask} One distinctive advantage of \textit{MetaAug} over approaches such as multi task learning is the ability to share knowledge across tasks while respecting task heterogeneity. 
    Specifically, 
    for each label space $\mathcal{Y}_i$, 
    we assume a trivially-casual mask, $\mathbf{CMask}(i) \subseteq \mathcal{C}^{meta}$, containing concepts with \textit{deterministic} causal connection to the label: 
    \begin{align}
        \mathbf{CMask}(i) \defeq \{x | (\vec{c}_s)_{(x)} \Rightarrow \mathbf{y}_i(s) \forall s \in \mathcal{S}_i, x \in \mathcal{C}^{meta} \} \label{eq-causal mask}
    \end{align}
    where $\Rightarrow$ stands for the logical implication. The casual mask must include the label itself,  $\mathcal{Y}_i \in \mathbf{CMask}(i)$, but may 
    cover other features. For example, if the label is ``$\text{age\_[0,18]}$" (for people of age between 0 to 18), the $CMask(i)$ also covers features like ``$\text{age\_[19,30]}$".
    As a result, the augmented concept vocabulary becomes 
    \begin{align} \label{eq:caug}
        \mathcal{C}_{Aug}^{i} \defeq \mathcal{C}^{meta} \setminus  \mathbf{CMask}(i)
    \end{align}
    where $\setminus$ denotes set difference operator.
    
    \noindent\textbf{Model architecture}
    In order to facilitate efficient optimization and adaptation, we propose a shared parameter space for meta-parameters
    $\Omega$ and task-specific parameters $\Theta_i$ for every single task
      that jointly models the conflicts and relations between tasks.
    To that end, we choose the mixture of experts model that captures 
    modularized information across different tasks \cite{jacobs1991adaptive} along with a gating \cite{ma2018modeling} mechanism, where tasks with low correlation and conflicts can be separated into different experts to reduce negative transfer \cite{zhang2020overcoming}.
    To that end, we divide the neural net into
    $K$ task specific prediction networks $\mathbf{Task}_i$,
    $E$ expert networks $\mathbf{Expert}_j$,
    and $K$ task specific gating networks $\mathbf{Gate}_i$.
    
    Consider the most general case, where an incoming data instance associated with an instance $s$ in $\mathcal{D}^{meta}$, with association to the $\mathcal{C}^{meta}$ described by  $\vec{c_s^{meta}}$.
    The $i$th output node of model, $h_i(\vec{c_s^{meta}})$, will correspond to the prediction for label $\mathcal{Y}_i$ of task $i$,  
    as follows (See \autoref{fig-network})
    \begin{align}
        h_i(\vec{c_s^{meta}}) & = \mathbf{Task}_i( (v_s)_i  ) 
    \end{align}    
    where $(v_s)_i$ is an element wise sum of expert networks output, each weighted by an individual component of $\mathbf{Gate}_i(\vec{c_s^{meta}})$ after normalizing into unit simplex via $\textrm{softmax}(\cdot)$, formally, 
    
    \begin{align}
        (v_s)_i &  = \sum_j^{E} \Big( \textrm{softmax} (\mathbf{Gate}_i(\vec{c_s^{meta}}))_{(j)} \cdot \mathbf{Expert}_j(\vec{c_s^{meta}}) \Big)
    \end{align}
    
    Finally, the discrepancy between individual task solution space $\mathcal{C}_{Aug}^i$ and global vocabulary $\mathcal{C}^{meta}$ is resolved using a simple mask trick: by setting corresponding entries in $\vec{c_s^{meta}}$ to zero and passing through the network, and taking corresponding head as shown in \autoref{fig-network}.
    Conversely, given an instance with task-specific vector with augmentation, we can pad corresponding entries with 0 and adapt it to the network, more detail about padding will be discussed in the coming section.

    \section{meta optimization}
    Many use cases of predictive segments require high reliability and efficiency of the learning system.
    In contrast to previous works that optimized meta parameters by 
    approximating their influence on downstream tasks 
    with a restrictive task specific adaptation procedure \cite{finn2017model, franceschi2018bilevel} or model class
    \cite{rajeswaran2019meta,snell2017prototypical}, 
    we develop a \textit{synchronous meta optimization} procedure for directly optimizing towards the end goal of every individual task \autoref{eq:con-meta}.
    By exploiting the close correspondence between meta learner architecture $\Omega$ and task-specific learner architecture $\Theta_i$,
     The high level intuition is to jointly train on every individual task with respect to the meta parameter instance
     $\omega$ and task-specific learner architecture $\theta_i$ with shared memory. 
     Our optimization method makes no assumption on architecture of single task learner, nor the specific class of tasks and objective, other than the fact that each task specific learner is parameterized by $\theta_i$
     and there exist gradients from corresponding loss function.     
     \jason{either "gradients from corresponding loss functions" or "gradients from the corresponding loss function"}
     
     \noindent\textbf{Meta-Loss Function} Formally, we propose an end-to-end learning procedure of the meta-learner for optimizing a meta loss function that closely resembles the end goal ( \autoref{eq:con-meta}), as 
     \begin{align} 
& \mathcal{L}^{meta}(
 \theta^{*~(i)}(\omega), \omega, \mathcal{D}_{source})\defeq \nonumber\\
&  \sum^{K}_{i=1} \left(
 \sum_{s \in \mathcal{D}^{train~(i)}_{source} }
\mathbf{L}_i(h_i(\vec{c_s}; \theta^{*~(i)}(\omega), \omega), \mathbf{y}_i(s))\right)
          \label{eq:loss}
     \end{align}

\noindent with the corresponding $\theta^{*~(i)}(\omega)$ being the task specific output. For each instance $s_i$ from task $i$ with concept vocabulary $\mathcal{C}_{Aug~i}$, 
     \begin{align}
\theta^{*~(i)}(\omega) \defeq h_i(\mathbf{PadMask}(\vec{c_{s_i}}) ) \label{eq:theta}         
     \end{align}

\noindent where $\mathbf{PadMask}$ denotes the process of padding concept vector $\vec{c_{s_i}}$ with 
    vocabulary $\mathcal{C}_{Aug}^i$ on to the vector with vocabulary $\mathcal{C}^{meta}$ with missing entry set as\jason(of?) zero as illustrated in \autoref{fig-network}. 
    
    \noindent \textbf{Mixed experience replay} To mitigate catastrophic forgetting, we propose an \textit{end to end} learning method for obtaining the optimal meta parameter $\omega^*$ that minimizes the meta objective in \autoref{eq:loss} in an end to end fashion, without resorting to second order gradients.
     
     Specifically, we follow an \textit{in-batch task mixing} procedure that learns 
     a meta-parameter that directly optimizes on each of the end tasks' loss without task specific adaptation.
     The key to the optimization lies in the  gradient update step.
     For each step we create a mini-batch of instances
     $\mathcal{D}^{batch}$
     , where each instance $s_i \in \mathcal{D}^{batch}$ corresponds to a task $\mathcal{T}^i$ with $i$ sampled\jason{samples?} in proportion to $\mathcal{D}^{train~(i)}_{source}$, with the loss associated with the instance dynamically computed according to the task $i$ it belongs to. The resulting gradient update to the meta-parameter $\omega$ is the sum over all task-specific instances in the batch

\vspace{-0.15cm}
    \begin{equation}
    \vspace{-0.2cm}
    \omega \leftarrow \omega - \eta \nabla_\omega \sum_{s_i \in \mathcal{D}^{batch}}  \sum_{j=1}^{K} 
    \Big (
    \mathbb{I}(i = j) \mathbf{L}_i(h_i(\vec{c_s}; \theta^{*~(i)}(\omega), \omega), \mathbf{y}_i(s))
    \Big) 
    \label{eq:metaupdate}
    \end{equation}
    where $s_i$ indicates instance for task $i$ sampled from $\mathcal{D}^{batch}$, 
    $\eta$ is the meta step size,
    and $\theta^{*~(i)}(\omega)$ is defined according to \autoref{eq:theta}.

We present the following performance guarantee based on the notion of \textit{Learner Advantage}
    
 \begin{Definition} \label{def-advantage} [Asymptotic Learner Advantage under $(\alpha,\beta)$ regret]\label{def hetero}
 We say there is a learner advantage of a model $f(\cdot;\theta)$ over model $g(\cdot;\phi)$ 
 if there exists \textit{constant}, $\alpha$, $\beta$, 
 so that for every input distribution $p(\cdot)$, label function $\mathbf{y}(z)$ and smooth loss function $\mathbf{L}(\cdot,\mathbf{y}(x))$,
 given enough data and optimization trivial\jason{I'm not sure what is being said here}, the resulted optimized model performance $f(\cdot;\theta^{*})$ and $g(\cdot;\phi^{*})$ satisfies 
 \begin{align}
     \mathbb{E}_{x \sim p(x)} (\mathbf{L}(f(x;\theta^{*}), \mathbf{y}(x)) ) \leq \mathbb{E}_{x \sim p(x)} (\mathbf{L}(g(x;\phi^{*}), \mathbf{y}(x)) )
 \end{align}
 with $f(\cdot;\theta^{*})$ consuming no more than 
 $\alpha $ times the number of parameters, 
 and $\beta$ times the number flops per model invocation,
 and $\beta$ times the number flops per model backward computation,
 than that of $g(\cdot;\phi^{*})$.
\end{Definition}

We have the following advantage results regarding arbitrary combinations of single task learners or multi-task learner architecture.
\begin{Theorem}
There exists learner advantage $(1+\mathcal{O}(1/n),1+\mathcal{O}(1/n))$ of \textit{MetaAug} 
$\{h_i(\vec{c_s}; \theta^{*~(i)}(\omega), \omega)\}_{i=1}^K$
over a combination of $K$ arbitrary single task learners (each applied to input instances of the corresponding task),
$\{g_i(\vec{c}_s; \phi_i)\}_{i=1}^K$,
and a $(1+\mathcal{O}(1/n))-(1+\mathcal{O}(1/n))$ learner advantage of \textit{MetaAug}  
over arbitrary multi-task learner 
$\vec{g}(\vec{c}_s; \phi)$ 
for the loss and label defined in Definition \ref{def-advantage},
where $n$ denotes the number of parameter and number of flops of computation.
\end{Theorem}
\noindent\textbf{Proof} The inequality can be constructed by considering intermediate $\tilde{\omega}$ such that 
$\mathbf{Expert}_i \defeq g_i( \mathbf{Pad}(\vec{c}_s;\mathcal{C}_{Aug~i}); \phi_i), 1 \leq i \leq K$
for single task learner combination cases, 
or 
$\mathbf{Expert}_0 \defeq \vec{g}(\mathbf{Pad}(\vec{c}_s; \mathcal{C}_{meta} \setminus \bigcup_{i=1}^K \mathbf{CMask}(i)); \phi)$ 
for the multi-task learner case, 
where 
$\mathbf{Pad}(\vec{c}_s;\mathcal{C})$ pads the $\vec{c}_s$ 
with corresponding values indicating associations between entity $s$ and $c$
along with $\mathbf{Task}_i(v) \defeq $Identity$(v)$
and properly defined $\{\mathbf{Gate}_i\}_{i=1}^K$, 
along with the following fact due to the loss function as defined in \autoref{eq:loss}
\begin{align} \label{eq-optimized-result}
     \mathbb{E}_{i \sim \mathbb{T}(i)} (
 \mathbb{E}_{s \sim \mathbb{P}_i(s)}
\mathbf{L}_i(h_i(\vec{c_s}; \theta^{*~(i)}(\omega^*), \omega^*), \mathbf{y}_i(s))
) \leq \nonumber\\
\mathbb{E}_{i \sim \mathbb{T}(i)} (
 \mathbb{E}_{s \sim \mathbb{P}_i(s)}
\mathbf{L}_i(h_i(\vec{c_s}; \theta^{*~(i)}(\tilde{\omega}), \tilde{\omega}), \mathbf{y}_i(s))
)
 \end{align}.

We also have the following result considering learner advantage over the family of MAML algorithms that optimize the meta parameter by adaptation performance (Equation 1 in \cite{finn2017model}).
\begin{Theorem}
There exists learner advantage $(1+\mathcal{O}(1/n),(1/K+1)+\mathcal{O}(1/n))$ of \textit{MetaAug}
$\{h_i(\vec{c_s}; \theta^{*~(i)}(\omega), \omega)\}_{i=1}^K$
and MAML algorithm $g_\phi(x)$ where $K$ denotes the number of adaptation samples and $n$ denotes the number of parameter and number of flops of computation.
\end{Theorem} 

\section{Extension to single-task learning}
\noindent\textbf{Online Adaptation}\label{sec:single task}
The original \textit{MetaCon} approach requires the presence of $K$ task-specific datasets as well as computation intensive joint training for the meta-dataset, which may hinder its ability to quickly adapt to incoming traffic data that are only available to a specific single task $\mathcal{T}_i$. 
To that end, we implement an Online Adaptation procedure.
Given an updated dataset 
$(\mathcal{D}^{train~'}, \mathcal{D}^{val~'}, \mathcal{D}^{test~'})^{(i)}$ 
and 
the trained meta-model $\omega^*$ as the result of Algorithm 1,
we perform standard gradient based optimization from initial state 
$\theta_o \defeq \theta^{*~(i)}(\omega)$ (See \autoref{eq:theta}) 
similar to the fine-tuning procedure from 
large scale pre-trained model \cite{tan2020tnt}.

\noindent\textbf{Single Task Meta Learning}
The original \textit{MetaCon} can also be extended to solve single task meta learning problems where there is only one task $\mathcal{T}_0$,
with entity set 
$\mathcal{S}$,
dataset 
$\mathcal{D}^{train}, \mathcal{D}^{val}, \mathcal{D}^{test}$,
label function
$\mathbf{y}: \mathcal{S} \to \mathcal{Y}$ 
and concept vocabulary $\mathcal{C}$, by following a masked concept learning procedure.
Specifically, 
we construct auxiliary tasks 
$\mathcal{T}_x$ for each $x \in \mathcal{C}$, 
with 
label $\mathbf{y}(s) \defeq (\Vec{c}_s)_{x}$,
casual mask as defined in \autoref{eq-causal mask},
$\mathcal{C}_{Aug x}$ as defined in \autoref{eq:caug}, and with 
$\mathcal{C}_{meta} \defeq \mathcal{C} \bigcup \{\mathcal{Y}\}$.
The Single Task Meta Learning can be implemented as first performing meta learning with respect to tasks 
$\{\mathcal{T}_0\} \big \{ \mathcal{T}_x | x \in \mathcal{C} \}$ optionally followed by the \textit{Online Adaptation} algorithm.
\begin{table*}[]
\centering
\resizebox{\textwidth}{!}{%
\begin{tabular}{|c|c|c|cc|cc|cc|cc|cc|cc|}
\hline
\rowcolor[HTML]{C0C0C0} 
 &
   &
  WDL &
  \multicolumn{2}{c|}{\cellcolor[HTML]{C0C0C0}PLE} &
  \multicolumn{2}{c|}{\cellcolor[HTML]{C0C0C0}MMOE} &
  \multicolumn{2}{c|}{\cellcolor[HTML]{C0C0C0}ESSM} &
  \multicolumn{2}{c|}{\cellcolor[HTML]{C0C0C0}DCNMix} &
  \multicolumn{2}{c|}{\cellcolor[HTML]{C0C0C0}DCN} &
  \multicolumn{2}{c|}{\cellcolor[HTML]{C0C0C0}MetaCon} \\ \hline
\rowcolor[HTML]{9B9B9B} 
{\color[HTML]{000000} } &
  {\color[HTML]{000000} } &
  {\color[HTML]{000000} Absolute} &
  \multicolumn{1}{c|}{\cellcolor[HTML]{9B9B9B}{\color[HTML]{000000} Absolute}} &
  {\color[HTML]{000000} Relative} &
  \multicolumn{1}{c|}{\cellcolor[HTML]{9B9B9B}{\color[HTML]{000000} Absolute}} &
  {\color[HTML]{000000} Relative} &
  \multicolumn{1}{c|}{\cellcolor[HTML]{9B9B9B}{\color[HTML]{000000} Absolute}} &
  {\color[HTML]{000000} Relative} &
  \multicolumn{1}{c|}{\cellcolor[HTML]{9B9B9B}{\color[HTML]{000000} Absolute}} &
  {\color[HTML]{000000} Relative} &
  \multicolumn{1}{c|}{\cellcolor[HTML]{9B9B9B}{\color[HTML]{000000} Absolute}} &
  Relative &
  \multicolumn{1}{c|}{\cellcolor[HTML]{9B9B9B}Absolute} &
  Relative \\ \hline\hline
 &
  Accuracy &
  0.8227 &
  \multicolumn{1}{c|}{0.8405} &
  +2.16\% &
  \multicolumn{1}{c|}{0.8413} &
  +2.26\% &
  \multicolumn{1}{c|}{0.8148} &
  -0.96\% &
  \multicolumn{1}{c|}{0.7985} &
  -2.94\% &
  \multicolumn{1}{c|}{0.8071} &
  -1.90\% &
  \multicolumn{1}{c|}{\textbf{0.8417}} &
  +2.31\% \\ \cline{2-15} 
 &
  AUC &
  0.8687 &
  \multicolumn{1}{c|}{0.8913} &
  +2.60\% &
  \multicolumn{1}{c|}{0.8938} &
  +2.89\% &
  \multicolumn{1}{c|}{0.8774} &
  +1.00\% &
  \multicolumn{1}{c|}{0.8433} &
  -2.92\% &
  \multicolumn{1}{c|}{0.8646} &
  -0.47\% &
  \multicolumn{1}{c|}{\textbf{0.894}} &
  +2.91\% \\ \cline{2-15} 
 &
  F1 &
  0.5122 &
  \multicolumn{1}{c|}{0.6165} &
  +20.36\% &
  \multicolumn{1}{c|}{0.6416} &
  +25.26\% &
  \multicolumn{1}{c|}{0.4289} &
  -16.26\% &
  \multicolumn{1}{c|}{0.3536} &
  -30.96\% &
  \multicolumn{1}{c|}{0.3957} &
  -22.75\% &
  \multicolumn{1}{c|}{\textbf{0.6505}} &
  +27.00\% \\ \cline{2-15} 
 &
  Kappa &
  0.4156 &
  \multicolumn{1}{c|}{0.5181} &
  +24.66\% &
  \multicolumn{1}{c|}{0.5403} &
  +30.00\% &
  \multicolumn{1}{c|}{0.3448} &
  -17.04\% &
  \multicolumn{1}{c|}{0.2701} &
  -35.01\% &
  \multicolumn{1}{c|}{0.3111} &
  -25.14\% &
  \multicolumn{1}{c|}{\textbf{0.5484}} &
  +31.95\% \\ \cline{2-15} 
 &
  Log loss &
  6.1224 &
  \multicolumn{1}{c|}{5.5094} &
  -10.01\% &
  \multicolumn{1}{c|}{5.4797} &
  -10.50\% &
  \multicolumn{1}{c|}{6.3961} &
  +4.47\% &
  \multicolumn{1}{c|}{6.9583} &
  +13.65\% &
  \multicolumn{1}{c|}{6.6613} &
  +8.80\% &
  \multicolumn{1}{c|}{\textbf{5.4691}} &
  -10.67\% \\ \cline{2-15} 
\multirow{-6}{*}{a9a} &
  Overall &
   &
  \multicolumn{1}{c|}{} &
  +59.80\% &
  \multicolumn{1}{c|}{} &
  +70.92\% &
  \multicolumn{1}{c|}{} &
  -37.73\% &
  \multicolumn{1}{c|}{} &
  -85.49\% &
  \multicolumn{1}{c|}{} &
  -59.06\% &
  \multicolumn{1}{c|}{} &
  +74.85\% \\ \hline\hline
\cellcolor[HTML]{FFFFFF}{\color[HTML]{222222} } &
  Accuracy &
  0.505 &
  \multicolumn{1}{c|}{0.5033} &
  -0.34\% &
  \multicolumn{1}{c|}{0.59} &
  +16.83\% &
  \multicolumn{1}{c|}{0.545} &
  +7.92\% &
  \multicolumn{1}{c|}{0.5950} &
  +17.82\% &
  \multicolumn{1}{c|}{0.5017} &
  -0.65\% &
  \multicolumn{1}{c|}{\textbf{0.6350}} &
  +25.74\% \\ \cline{2-15} 
\cellcolor[HTML]{FFFFFF}{\color[HTML]{222222} } &
  AUC &
  0.5162 &
  \multicolumn{1}{c|}{0.5083} &
  -1.53\% &
  \multicolumn{1}{c|}{0.6234} &
  +20.77\% &
  \multicolumn{1}{c|}{0.5627} &
  +9.01\% &
  \multicolumn{1}{c|}{0.6101} &
  +18.19\% &
  \multicolumn{1}{c|}{0.5017} &
  -2.81\% &
  \multicolumn{1}{c|}{\textbf{0.6841}} &
  +32.53\% \\ \cline{2-15} 
\cellcolor[HTML]{FFFFFF}{\color[HTML]{222222} } &
  F1 &
  0.5139 &
  \multicolumn{1}{c|}{0.6005} &
  +16.85\% &
  \multicolumn{1}{c|}{0.59} &
  +14.81\% &
  \multicolumn{1}{c|}{0.5269} &
  +2.53\% &
  \multicolumn{1}{c|}{0.5744} &
  +11.77\% &
  \multicolumn{1}{c|}{\textbf{0.6659}} &
  +29.58\% &
  \multicolumn{1}{c|}{0.6485} &
  +26.19\% \\ \cline{2-15} 
\cellcolor[HTML]{FFFFFF}{\color[HTML]{222222} } &
  Kappa &
  0.01 &
  \multicolumn{1}{c|}{0.0067} &
  -33.00\% &
  \multicolumn{1}{c|}{0.18} &
  +1700.00\% &
  \multicolumn{1}{c|}{0.09} &
  +800.00\% &
  \multicolumn{1}{c|}{0.1900} &
  +1800.00\% &
  \multicolumn{1}{c|}{0.0033} &
  -67.00\% &
  \multicolumn{1}{c|}{\textbf{0.2700}} &
  +2600.00\% \\ \cline{2-15} 
\cellcolor[HTML]{FFFFFF}{\color[HTML]{222222} } &
  Log loss &
  17.0969 &
  \multicolumn{1}{c|}{17.1546} &
  +0.34\% &
  \multicolumn{1}{c|}{14.1611} &
  -17.17\% &
  \multicolumn{1}{c|}{15.7153} &
  -8.08\% &
  \multicolumn{1}{c|}{13.9883} &
  -18.18\% &
  \multicolumn{1}{c|}{17.2122} &
  +0.67\% &
  \multicolumn{1}{c|}{\textbf{12.6068}} &
  -26.26\% \\ \cline{2-15} 
\multirow{-6}{*}{\cellcolor[HTML]{FFFFFF}{\color[HTML]{222222} madelon}} &
  Overall &
   &
  \multicolumn{1}{c|}{} &
  -18.35\% &
  \multicolumn{1}{c|}{} &
  +1769.58\% &
  \multicolumn{1}{c|}{} &
  +827.54\% &
  \multicolumn{1}{c|}{} &
  +1865.97\% &
  \multicolumn{1}{c|}{} &
  -41.56\% &
  \multicolumn{1}{c|}{} &
  +2710.72\% \\ \hline
\end{tabular}%
}
\caption{Overall performance comparison among MetaCon and baseline methods on the public datasets \texttt{A9A} and \texttt{Madelon}. 
The classification metric AUC, Accuracy, F1 score, 
and the ranking metric Kappa, and Log loss 
against the ground truth according to both the absolute value and relative value compared to baseline. 
}
\end{table*}

\begin{figure*}[]
\centering
\vspace{-10pt}
\includegraphics[width=\textwidth]{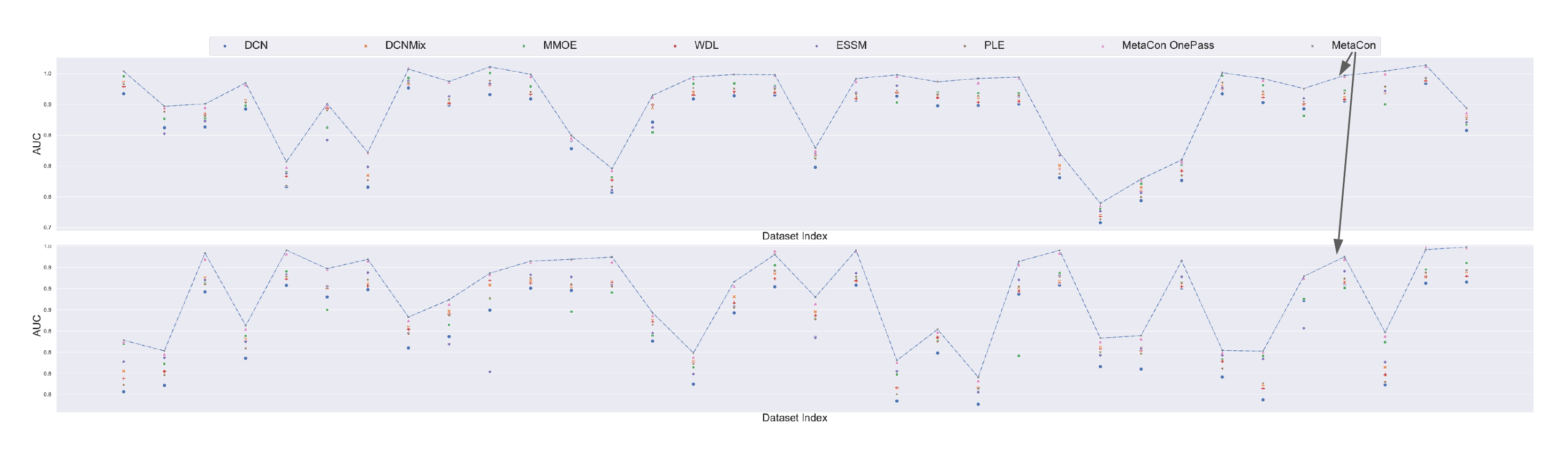}
\caption{Core performance comparison over the 68 production predictive segment tasks \yifan{without legend, the reader can't understand what each line is...please add call-out in the figure. Mention that there are 68 tasks, corresponding to x-axis, each line represent a method}}
\label{fig-AUC}
\end{figure*}

\begin{figure*}[]
\centering
\includegraphics[width=\textwidth]{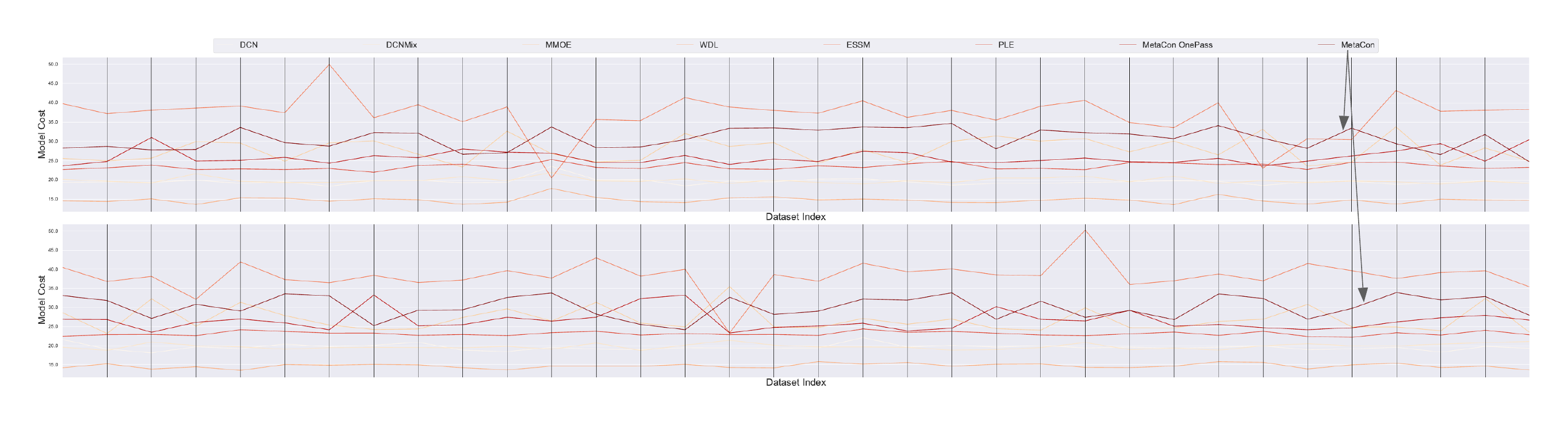}
\caption{Model cost measured in milliseconds over the 68 production predictive segment tasks \yifan{without legend, the reader can't  understand what each line is...please add call out. Also make the color scheme of 5 and 6 consistent}}
\label{fig-inference cost}
\end{figure*}

\begin{figure}[]
\centering
\includegraphics[width=.5\textwidth]{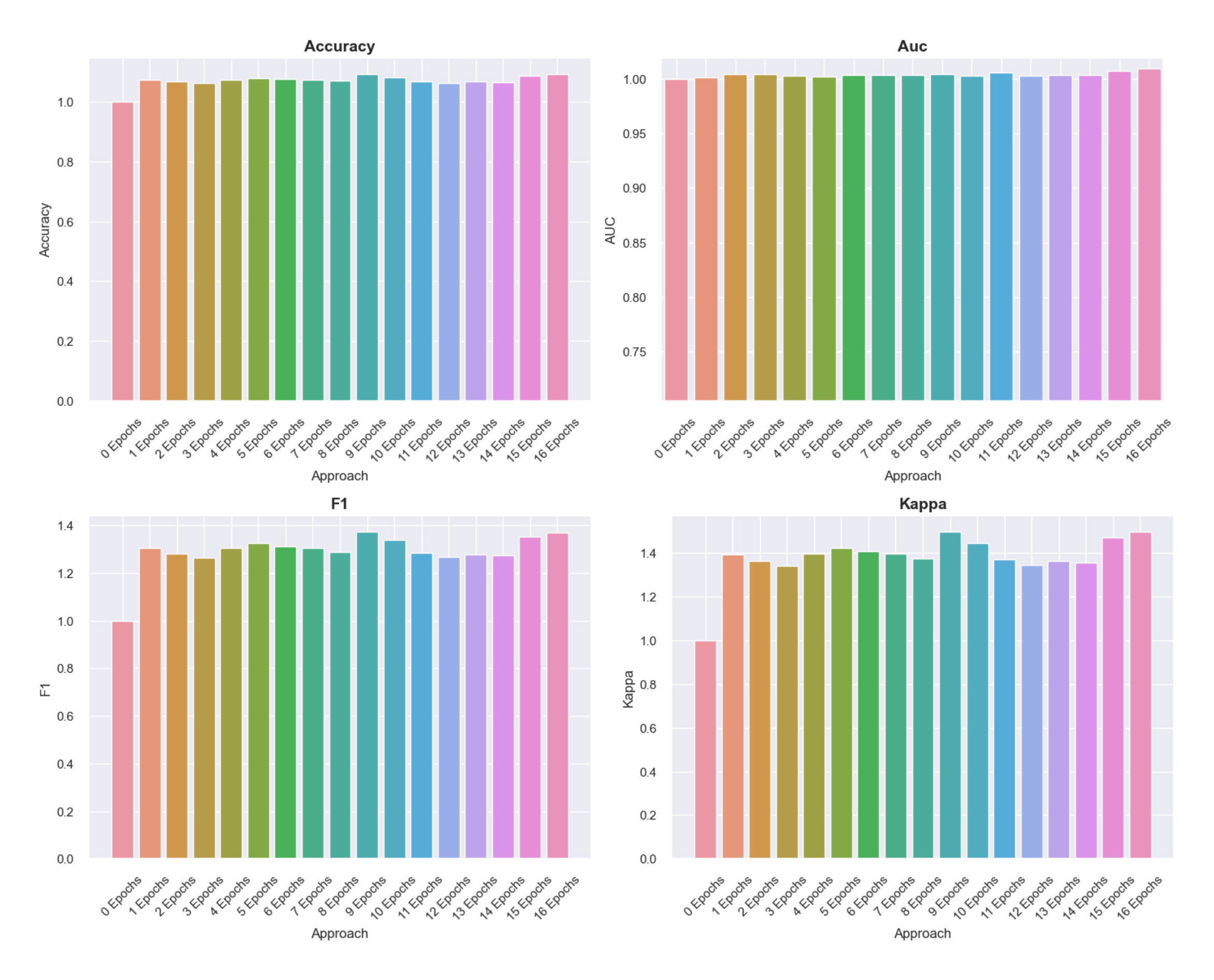}

\caption{Performance comparison of \textit{MetaCon} variants 
with different Meta-train Epochs }
\label{fig-Quantitative}

\end{figure}

\begin{figure}[]
\centering
\includegraphics[width=.5\textwidth]{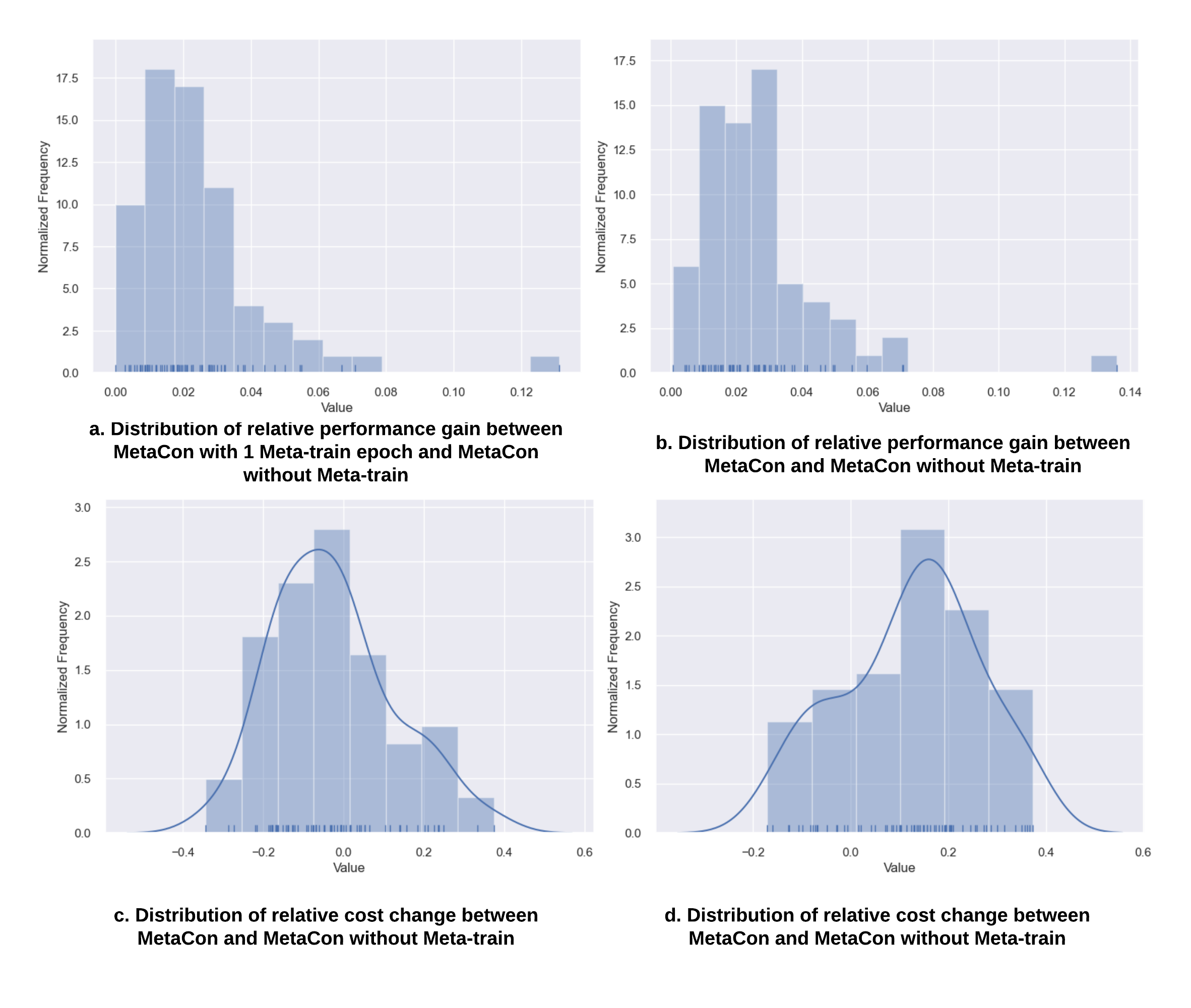}
\caption{Distribution of change in performance and model cost 
across all predictive segment tasks between \textit{MetaCon} variants.
}
\label{fig-qualitative}
\end{figure}

\begin{figure}[]
\centering
\includegraphics[width=.45\textwidth]{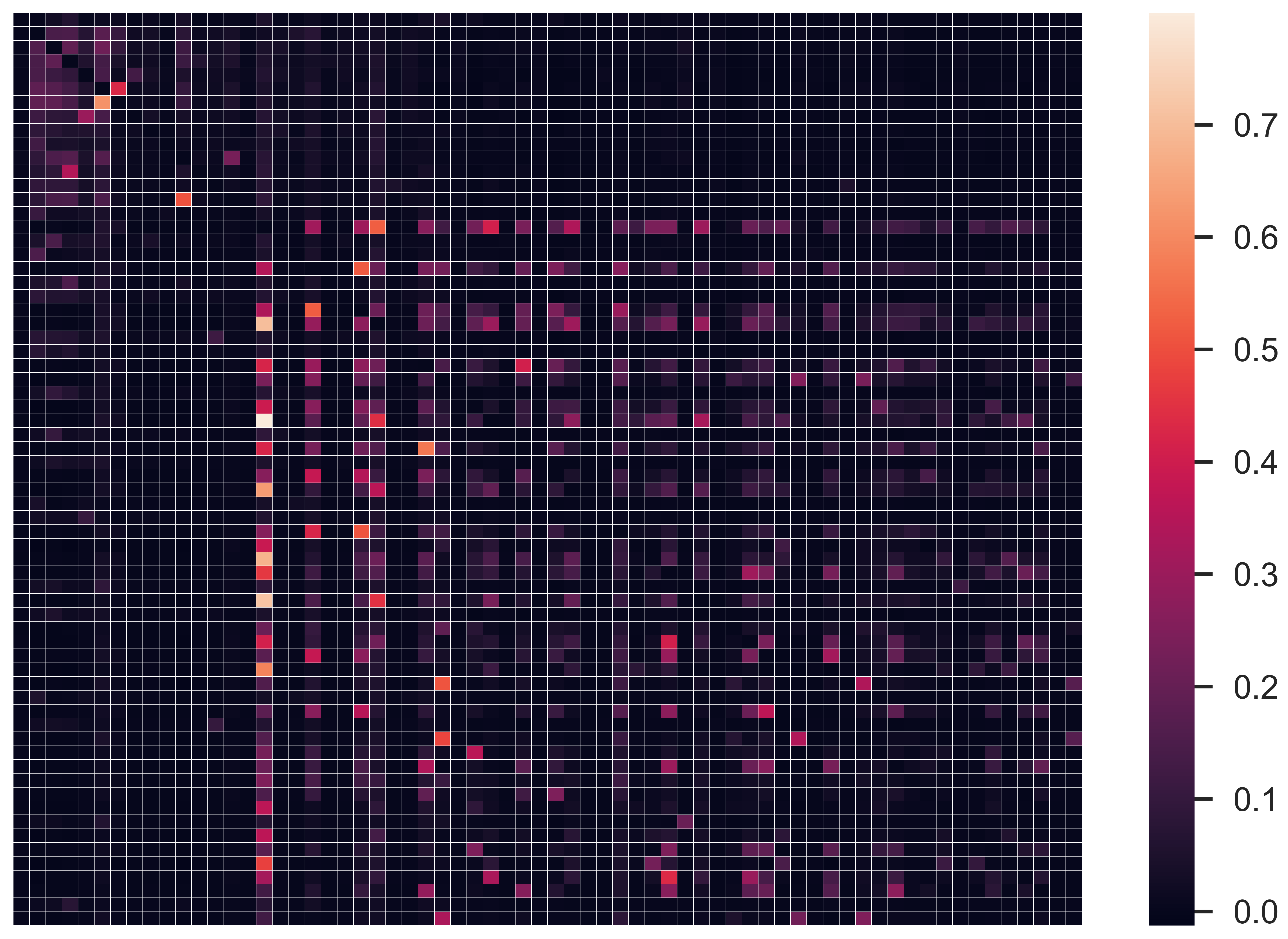}
\caption{Task attention learned by \textit{Metacon} across the 68 predictive segment tasks}
\label{fig-Interpret-ability study on task relation learning}
\vspace{-20pt}
\end{figure}

\section{EXPERIMENT}
In this section, we present a series of experiments centered around the following research questions: 
\begin{itemize}[leftmargin=2.5em]
    \item[\bf RQ1]  How does the performance of \textit{MetaCon} compare to alternative approaches for key industrial use cases?
    \item[\bf RQ2]  How do the settings and individual components of \textit{MetaCon} impact its performance quantitatively and qualitatively? 
    \item[\bf RQ3]  How does the computation effort for the meta-training impact downstream results?
    \item[\bf RQ4] How reliable and interpretable is \textit{MetaCon} in meeting human intuition?
    \item[\bf RQ5] How does \textit{MetaCon} compare to baselines of effectiveness-efficiency tradeoff and computation cost?
    \item[\bf RQ6] What degree of generalizability does single task meta learning approach of  \textit{MetaCon} achieve on public structured data learning tasks?
    \item[\bf RQ7] How robust is the performance of \textit{MetaCon} compared with alternatives in key production systems? 
\end{itemize}

\noindent\textbf{Dataset}
We use both proprietary and public datasets. For the former, we collected and compiled a total of 68 different segment prediction tasks 
in production regarding 
users' association with interest taxonomy including 
\textbf{}Yahoo Content Taxonomy, Oath interest Category similar to \cite{zhou2016predicting}, 
 as well as Wikipedia and Price-Grabber. This 
 dataset has a 100K dimensional unfolded vector per instance, and a total of 100K instances. We split the dataset \jason{into?} 3 folds, with 2/3 being the support set and 1/3 being the hold-out query/test set.
 
We extend our methodology in Section 6 and apply it to the single task learning scenario on public structured learning datasets.
Specifically, we use the 
 \madelon \cite{guyon2004result} 
 and 
 \ana \cite{platt1998sequential}, 
 which contain 2,000 training samples, 600 test samples with 500 features per sample, and 
 32,561 training samples, 16,281 test samples with 123 features per sample, respectively.

\noindent\textbf{Compared Methods}
We implement \textit{MetaCon} in two variants:
a \textit{OnePass} variant that trains the meta parameter $\omega$ over one pass of the full 68 tasks along with the additional 1000 auxiliary tasks constructed according to the method in Section 6, which is then optimized with meta-testing stage with single task adaptation (\autoref{sec:single task});
and one that trains with 15 passes, resulting in a total computation of 100 trillion unfolded concepts. 
Each variant of \textit{MetaCon} uses 3 \textit{experts} (See \autoref{fig-network}), with each expert neural network having 6 repeated layers of a width of 512 with dense residual connections, 
each task specific gating networks having a single hidden layer size of 32,
and each task specific networks with a single hidden layer size of 32  (See \autoref{fig-network}).
\jason{are the last two lines saying the same thing?}

In addition, we implement the following baseline approaches:
\begin{itemize}

\item \textbf{PLE} implements Progressive Layered Extraction \cite{tang2020progressive} with 1 shared expert and 2 specific experts, an expert layer width of 256, 256, gate layer width of 16, 16, tower layer depth of 32, 32.
\item \textbf{WDL} implements wide and deep learning  \cite{cheng2016wide} with deep part having layer width of 8 for \madelon and 256, 128, 64 for the rest dataset.
\item \textbf{MMOE} implements multi-gated mixture of experts \cite{zhao2019recommending} with 3 experts each with 6 repeated layer of a width of 512 as in the \textit{MetaCon} setting.
\item \textbf{ESSM}  implements Entire Space Multi-Task Model \cite{ma2018entire} with CTR and CVR each with 
layer width 512, 512, and the rest  following original defaults.
\item \textbf{DCN} implements Deep \& Cross Network \cite{wang2017deep} with layer width 384, 128, 64, cross number of 2 and cross dimension of 100 with rest of setting following original default.
\item \textbf{DCNMix}  implements Cost-Effective Mixture of Low-Rank DCN \cite{wang2021dcn} with 4 expert each with layer width 256, 128, 64, 
cross number of 2 and cross dimension of 100 with rank of 32.
\end{itemize}

All online adaptation and single task learning was performed with 30 epochs of Adam optimization with a tuned learning rate between 1e-6 and 1e-5.

\noindent\textbf{Core Performance Evaluation [RQ1] }
First, we compare the performance of the \textit{MetaCon} against baseline  approaches 
over the 68 production predictive segments tasks and 
score their performance in the core production metric of ROC-AUC.
As shown in \autoref{fig-AUC}, \textit{MetaCon} consistently outperforms candidate baselines 
across the 68 tasks 
with a clear distinction between its \textit{OnePass} version, followed by other strong baselines including \textit{MMOE}, \textit{PLE} and \textit{ESSM}.

\noindent\textbf{Computation Cost [RQ5] }
For deployment to production system, 
it is critical to balance between model performance and its computation cost. 
Here,
we study the computational cost in a per mapper node setting, 
where sharded subsets of data are sent to the local node to process sequentially. 
\autoref{fig-inference cost} 
shows the per instance latency in the unit of micro-seconds, 
demonstrating that different version of \textit{MetaCon} fares well in the computation cost due to its relative parallel architecture.

\noindent\textbf{ Ablation Analysis [RQ2-3]}
We first perform a qualitative study on the impact of meta training over downstream scoring metrics. 
The left part of figure \autoref{fig-qualitative} shows the distribution over the 68 production predictive segment tasks of the relative change in performance as measured by ROC-AUC
(top figure) and cost as measured in the per mapper node setting above (bottom figure)
between the \textit{OnePass} variant, as compared to ones without meta training.
The right parts shows distribution of difference between the original version the the ones without meta training. \jason{what is "ones" referring to}
From the result, 
we can observe an uniform improvement in performance score without significant change in computation cost, with ones with full meta-training \jason{what is ones referring to} achieving the biggest gains.
 
We next compare the impact of meta training settings quantitatively. 
\autoref{fig-Quantitative} shows the performance on production demographics segment prediction task 
in terms of Accuracy, ROC-AUC, F1 Score and Cohen-Kappa score for variants of \textit{MetaCon} with different degree of meta training 
as measured by the number of meta-training epochs.
From the results,
we can observe a significant correlation between different degree of meta-training, 
with the performance quickly increase from 0 epochs to 1, and then gradually increase 
from lower degree of meta-training
to higher  degree.

\noindent\textbf{Interpret-ability Study [RQ4]}
We next study the interpret-ability of \textit{MetaCon} by visualizing the meta-learned task attention,
where the score from task $i$ to task $j$ is obtained by comparing the difference in the predicted log like-hood of the ground truth label $\mathbf{y}_i$ before and after the masking of $\mathbf{CMask}(j)$. \autoref{fig-Interpret-ability study on task relation learning} which shows the 4628 attention scores across the 68  predictive segment tasks, with an off-diagonal distribution of high relatedness scores where clustering patterns and mutually beneficial task attentions naturally emerge.

\noindent\textbf{Application to structured data learning  [RQ6]}
We further evaluate the generalizability of \textit{MetaCon} by applying it to single task meta learning and compare its absolute and relative performance  with the baseline of wide and deep learning (\textit{WDL}) along with other candidate approaches.
Table 1 shows their performance 
over the metrics of Accuracy, ROC-AUC (AUC in table), Cohen-Kappa Score (Kappa in table), F1 Score (F1 in table) and Log loss, 
as well as the overall performance score (Overall in table) 
computed
as signed improvement across all performance metrics, which demonstrates the superior performance of \textit{MetaCon}.

\noindent\textbf{Online evaluation [RQ7]}
The \textit{MetaCon} is rolled out to production targeting use cases in the internal Hadoop based deployment environment.
Evaluation in key AUC-based targeting task between our system and the existing production system
demonstrate significant gains in the tradeoff between false negative and false positive with change in ROC-AUC from 0.78  to 0.90, a 
15.4\% improvement.

    
\section{Conclusion}

In this work, we present \textit{MetaCon} as our unified predicative system with trillion concepts meta learning. It is built on top of a  unfolded concept representation framework, that utilizes
 user's heterogeneous digital footprint, to 
jointly learn over the entire spectrum of predicative tasks. 
Extensive evaluation on large number of predicative segment tasks and public benchmarks demonstrate the superior performance of  \textit{MetaCon} over state of the art recommendation and ranking approaches 
as well as the previous production system.

For future research, one particular interesting directions is to extend the \textit{MetaCon} paradigm for  specific modality and domain such as image and videos. In addition, deeper integration from commonsense knowledge is another promising direction to explore.
\yifan{The future direction is not clearly stated. When you say modality, given an example? e.g., extend the model to beyond text and into images and videos?}

\balance

{\footnotesize
\bibliographystyle{IEEEtran}
\bibliography{_main}

\begin{thebibliography}{10}
\providecommand{\url}[1]{#1}
\csname url@samestyle\endcsname
\providecommand{\newblock}{\relax}
\providecommand{\bibinfo}[2]{#2}
\providecommand{\BIBentrySTDinterwordspacing}{\spaceskip=0pt\relax}
\providecommand{\BIBentryALTinterwordstretchfactor}{4}
\providecommand{\BIBentryALTinterwordspacing}{\spaceskip=\fontdimen2\font plus
\BIBentryALTinterwordstretchfactor\fontdimen3\font minus
  \fontdimen4\font\relax}
\providecommand{\BIBforeignlanguage}[2]{{%
\expandafter\ifx\csname l@#1\endcsname\relax
\typeout{** WARNING: IEEEtran.bst: No hyphenation pattern has been}%
\typeout{** loaded for the language `#1'. Using the pattern for}%
\typeout{** the default language instead.}%
\else
\language=\csname l@#1\endcsname
\fi
#2}}
\providecommand{\BIBdecl}{\relax}
\BIBdecl

\bibitem{li2021hadoop}
K.~Li, Y.~Hu, M.~Verma, F.~Tan, C.~Hu, T.~Kasturi, and K.~Yen, ``Hadoop-mta: a
  system for multi data-center trillion concepts auto-ml atop hadoop,'' in
  \emph{2021 IEEE International Conference on Big Data (Big Data)}.\hskip 1em
  plus 0.5em minus 0.4em\relax IEEE Computer Society, 2021, pp. 5953--5955.

\bibitem{balusamy2021driving}
B.~Balusamy, S.~Kadry, A.~H. Gandomi \emph{et~al.}, ``Driving big data with
  hadoop tools and technologies,'' 2021.

\bibitem{cahill1997target}
D.~J. Cahill, ``Target marketing and segmentation: valid and useful tools for
  marketing,'' \emph{Management Decision}, 1997.

\bibitem{voigt2017eu}
P.~Voigt and A.~Von~dem Bussche, ``The eu general data protection regulation
  (gdpr),'' \emph{A Practical Guide, 1st Ed., Cham: Springer International
  Publishing}, vol.~10, p. 3152676, 2017.

\bibitem{chromium}
\url{https://blog.chromium.org/2019/05/improving-privacy-and-security-on-web.html},
  2019, [Online; accessed July-2019].

\bibitem{iqbal2021towards}
U.~Iqbal, ``Towards a privacy-preserving web,'' Ph.D. dissertation, The
  University of Iowa, 2021.

\bibitem{zhao2019recommending}
Z.~Zhao, L.~Hong, L.~Wei, J.~Chen, A.~Nath, S.~Andrews, A.~Kumthekar,
  M.~Sathiamoorthy, X.~Yi, and E.~Chi, ``Recommending what video to watch next:
  a multitask ranking system,'' in \emph{Proceedings of the 13th ACM Conference
  on Recommender Systems}, 2019, pp. 43--51.

\bibitem{mcmahan2013ad}
H.~B. McMahan, G.~Holt, D.~Sculley, M.~Young, D.~Ebner, J.~Grady, L.~Nie,
  T.~Phillips, E.~Davydov, D.~Golovin \emph{et~al.}, ``Ad click prediction: a
  view from the trenches,'' in \emph{Proceedings of the 19th ACM SIGKDD
  international conference on Knowledge discovery and data mining}, 2013, pp.
  1222--1230.

\bibitem{liu2017related}
D.~C. Liu, S.~Rogers, R.~Shiau, D.~Kislyuk, K.~C. Ma, Z.~Zhong, J.~Liu, and
  Y.~Jing, ``Related pins at pinterest: The evolution of a real-world
  recommender system,'' in \emph{Proceedings of the 26th international
  conference on world wide web companion}, 2017, pp. 583--592.

\bibitem{he2014practical}
X.~He, J.~Pan, O.~Jin, T.~Xu, B.~Liu, T.~Xu, Y.~Shi, A.~Atallah, R.~Herbrich,
  S.~Bowers \emph{et~al.}, ``Practical lessons from predicting clicks on ads at
  facebook,'' in \emph{Proceedings of the Eighth International Workshop on Data
  Mining for Online Advertising}, 2014, pp. 1--9.

\bibitem{zhai2017visual}
A.~Zhai, D.~Kislyuk, Y.~Jing, M.~Feng, E.~Tzeng, J.~Donahue, Y.~L. Du, and
  T.~Darrell, ``Visual discovery at pinterest,'' in \emph{Proceedings of the
  26th International Conference on World Wide Web Companion}, 2017, pp.
  515--524.

\bibitem{freno2017practical}
A.~Freno, ``Practical lessons from developing a large-scale recommender system
  at zalando,'' in \emph{Proceedings of the Eleventh ACM Conference on
  Recommender Systems}, 2017, pp. 251--259.

\bibitem{covington2016deep}
P.~Covington, J.~Adams, and E.~Sargin, ``Deep neural networks for youtube
  recommendations,'' in \emph{Proceedings of the 10th ACM conference on
  recommender systems}, 2016, pp. 191--198.

\bibitem{lu2018like}
Y.~Lu, R.~Dong, and B.~Smyth, ``Why i like it: multi-task learning for
  recommendation and explanation,'' in \emph{Proceedings of the 12th ACM
  Conference on Recommender Systems}, 2018, pp. 4--12.

\bibitem{agarwal2011localized}
D.~Agarwal, B.-C. Chen, and B.~Long, ``Localized factor models for
  multi-context recommendation,'' in \emph{Proceedings of the 17th ACM SIGKDD
  international conference on Knowledge discovery and data mining}, 2011, pp.
  609--617.

\bibitem{wang2016multi}
S.~Wang, M.~Gong, H.~Li, and J.~Yang, ``Multi-objective optimization for long
  tail recommendation,'' \emph{Knowledge-Based Systems}, vol. 104, pp.
  145--155, 2016.

\bibitem{ma2018entire}
X.~Ma, L.~Zhao, G.~Huang, Z.~Wang, Z.~Hu, X.~Zhu, and K.~Gai, ``Entire space
  multi-task model: An effective approach for estimating post-click conversion
  rate,'' in \emph{The 41st International ACM SIGIR Conference on Research \&
  Development in Information Retrieval}, 2018, pp. 1137--1140.

\bibitem{tang2020progressive}
H.~Tang, J.~Liu, M.~Zhao, and X.~Gong, ``Progressive layered extraction (ple):
  A novel multi-task learning (mtl) model for personalized recommendations,''
  in \emph{Fourteenth ACM Conference on Recommender Systems}, 2020, pp.
  269--278.

\bibitem{fischbein1996psychological}
E.~Fischbein, ``The psychological nature of concepts,'' in \emph{Mathematics
  for tomorrow’s young children}.\hskip 1em plus 0.5em minus 0.4em\relax
  Springer, 1996, pp. 102--119.

\bibitem{li2019mining}
K.~Li, ``Mining and analyzing technical knowledge based on concepts,'' Ph.D.
  dissertation, University of California Santa Barbara, 2019.

\bibitem{wang2015concept}
C.~Wang, K.~Chakrabarti, Y.~He, K.~Ganjam, Z.~Chen, and P.~A. Bernstein,
  ``Concept expansion using web tables,'' in \emph{WWW}.\hskip 1em plus 0.5em
  minus 0.4em\relax International World Wide Web Conferences Steering
  Committee, 2015, pp. 1198--1208.

\bibitem{li2014social}
K.~Li, W.~Lu, S.~Bhagat, L.~V. Lakshmanan, and C.~Yu, ``On social event
  organization,'' in \emph{Proceedings of the 20th ACM SIGKDD international
  conference on Knowledge discovery and data mining}, 2014, pp. 1206--1215.

\bibitem{li2017discovering}
K.~Li, Y.~He, and K.~Ganjam, ``Discovering enterprise concepts using
  spreadsheet tables,'' in \emph{Proceedings of the 23rd ACM SIGKDD
  International Conference on Knowledge Discovery and Data Mining}, 2017, pp.
  1873--1882.

\bibitem{zha2018fts}
H.~Zha, J.~Shen, K.~Li, W.~Greiff, M.~T. Vanni, J.~Han, and X.~Yan, ``Fts:
  Faceted taxonomy construction and search for scientific publications,'' 2018.

\bibitem{li2018poqaa}
K.~Li, P.~Zhang, H.~Liu, H.~Zha, and X.~Yan, ``Poqaa: Text mining and knowledge
  sharing for scientific publications,'' 2018.

\bibitem{zha2019mining}
H.~Zha, W.~Chen, K.~Li, and X.~Yan, ``Mining algorithm roadmap in scientific
  publications,'' in \emph{Proceedings of the 25th ACM SIGKDD International
  Conference on Knowledge Discovery \& Data Mining}, 2019, pp. 1083--1092.

\bibitem{li2019hiercon}
K.~Li, S.~Li, S.~Yavuz, H.~Zha, Y.~Su, and X.~Yan, ``Hiercon: Hierarchical
  organization of technical documents based on concepts,'' in \emph{2019 IEEE
  International Conference on Data Mining (ICDM)}.\hskip 1em plus 0.5em minus
  0.4em\relax IEEE, 2019, pp. 379--388.

\bibitem{li2018unsupervised}
K.~Li and et. al., ``Unsupervised neural categorization for scientific
  publications,'' in \emph{Proceedings of the 2018 SIAM International
  Conference on Data Mining}.\hskip 1em plus 0.5em minus 0.4em\relax SIAM,
  2018, pp. 37--45.

\bibitem{li2018concept}
K.~Li, H.~Zha, Y.~Su, and X.~Yan, ``Concept mining via embedding,'' in
  \emph{2018 IEEE International Conference on Data Mining (ICDM)}.\hskip 1em
  plus 0.5em minus 0.4em\relax IEEE, 2018, pp. 267--276.

\bibitem{thrun1998learning}
S.~Thrun and L.~Pratt, ``Learning to learn: Introduction and overview,'' in
  \emph{Learning to learn}.\hskip 1em plus 0.5em minus 0.4em\relax Springer,
  1998, pp. 3--17.

\bibitem{kirsch2019improving}
L.~Kirsch, S.~van Steenkiste, and J.~Schmidhuber, ``Improving generalization in
  meta reinforcement learning using learned objectives,'' \emph{arXiv preprint
  arXiv:1910.04098}, 2019.

\bibitem{schweighofer2003meta}
N.~Schweighofer and K.~Doya, ``Meta-learning in reinforcement learning,''
  \emph{Neural Networks}, vol.~16, no.~1, pp. 5--9, 2003.

\bibitem{franceschi2018bilevel}
L.~Franceschi, P.~Frasconi, S.~Salzo, R.~Grazzi, and M.~Pontil, ``Bilevel
  programming for hyperparameter optimization and meta-learning,'' in
  \emph{International Conference on Machine Learning}.\hskip 1em plus 0.5em
  minus 0.4em\relax PMLR, 2018, pp. 1568--1577.

\bibitem{cubuk2018autoaugment}
E.~D. Cubuk, B.~Zoph, D.~Mane, V.~Vasudevan, and Q.~V. Le, ``Autoaugment:
  Learning augmentation policies from data,'' \emph{arXiv preprint
  arXiv:1805.09501}, 2018.

\bibitem{zhou2020online}
W.~Zhou, Y.~Li, Y.~Yang, H.~Wang, and T.~M. Hospedales, ``Online meta-critic
  learning for off-policy actor-critic methods,'' \emph{arXiv preprint
  arXiv:2003.05334}, 2020.

\bibitem{bechtle2021meta}
S.~Bechtle, A.~Molchanov, Y.~Chebotar, E.~Grefenstette, L.~Righetti,
  G.~Sukhatme, and F.~Meier, ``Meta learning via learned loss,'' in \emph{2020
  25th International Conference on Pattern Recognition (ICPR)}.\hskip 1em plus
  0.5em minus 0.4em\relax IEEE, 2021, pp. 4161--4168.

\bibitem{liu2018darts}
H.~Liu, K.~Simonyan, and Y.~Yang, ``Darts: Differentiable architecture
  search,'' \emph{arXiv preprint arXiv:1806.09055}, 2018.

\bibitem{finn2017model}
C.~Finn, P.~Abbeel, and S.~Levine, ``Model-agnostic meta-learning for fast
  adaptation of deep networks,'' in \emph{International Conference on Machine
  Learning}.\hskip 1em plus 0.5em minus 0.4em\relax PMLR, 2017, pp. 1126--1135.

\bibitem{bengio1990learning}
Y.~Bengio, S.~Bengio, and J.~Cloutier, \emph{Learning a synaptic learning
  rule}.\hskip 1em plus 0.5em minus 0.4em\relax Citeseer, 1990.

\bibitem{li2021automl}
Y.~Li~et.al., ``Automl: From methodology to application,'' in \emph{Proceedings
  of the 30th ACM International Conference on Information \& Knowledge
  Management}, 2021, pp. 4853--4856.

\bibitem{yang2020heterogeneous}
C.~Yang, Y.~Xiao, Y.~Zhang, Y.~Sun, and J.~Han, ``Heterogeneous network
  representation learning: A unified framework with survey and benchmark,''
  \emph{IEEE Transactions on Knowledge and Data Engineering}, 2020.

\bibitem{bayer2009evolving}
J.~Bayer, D.~Wierstra, J.~Togelius, and J.~Schmidhuber, ``Evolving memory cell
  structures for sequence learning,'' in \emph{International Conference on
  Artificial Neural Networks}.\hskip 1em plus 0.5em minus 0.4em\relax Springer,
  2009, pp. 755--764.

\bibitem{zoph2016neural}
B.~Zoph and Q.~V. Le, ``Neural architecture search with reinforcement
  learning,'' \emph{arXiv preprint arXiv:1611.01578}, 2016.

\bibitem{metz2018meta}
L.~Metz, N.~Maheswaranathan, B.~Cheung, and J.~Sohl-Dickstein, ``Meta-learning
  update rules for unsupervised representation learning,'' \emph{arXiv preprint
  arXiv:1804.00222}, 2018.

\bibitem{salimans2017evolution}
T.~Salimans, J.~Ho, X.~Chen, S.~Sidor, and I.~Sutskever, ``Evolution strategies
  as a scalable alternative to reinforcement learning,'' \emph{arXiv preprint
  arXiv:1703.03864}, 2017.

\bibitem{ravi2016optimization}
S.~Ravi and H.~Larochelle, ``Optimization as a model for few-shot learning,''
  2016.

\bibitem{snell2017prototypical}
J.~Snell, K.~Swersky, and R.~S. Zemel, ``Prototypical networks for few-shot
  learning,'' \emph{arXiv preprint arXiv:1703.05175}, 2017.

\bibitem{codd2002relational}
E.~F. Codd, ``A relational model of data for large shared data banks,'' in
  \emph{Software pioneers}.\hskip 1em plus 0.5em minus 0.4em\relax Springer,
  2002, pp. 263--294.

\bibitem{jacobs1991adaptive}
R.~A. Jacobs, M.~I. Jordan, S.~J. Nowlan, and G.~E. Hinton, ``Adaptive mixtures
  of local experts,'' \emph{Neural computation}, vol.~3, no.~1, pp. 79--87,
  1991.

\bibitem{ma2018modeling}
J.~Ma, Z.~Zhao, X.~Yi, J.~Chen, L.~Hong, and E.~H. Chi, ``Modeling task
  relationships in multi-task learning with multi-gate mixture-of-experts,'' in
  \emph{Proceedings of the 24th ACM SIGKDD International Conference on
  Knowledge Discovery \& Data Mining}, 2018, pp. 1930--1939.

\bibitem{zhang2020overcoming}
W.~Zhang, L.~Deng, L.~Zhang, and D.~Wu, ``Overcoming negative transfer: A
  survey,'' \emph{arXiv preprint arXiv:2009.00909}, 2020.

\bibitem{rajeswaran2019meta}
A.~Rajeswaran, C.~Finn, S.~Kakade, and S.~Levine, ``Meta-learning with implicit
  gradients,'' 2019.

\bibitem{tan2020tnt}
F.~Tan, Y.~Hu, C.~Hu, K.~Li, and K.~Yen, ``Tnt: Text normalization based
  pre-training of transformers for content moderation,'' in \emph{Proceedings
  of the 2020 Conference on Empirical Methods in Natural Language Processing
  (EMNLP)}, 2020, pp. 4735--4741.

\bibitem{zhou2016predicting}
K.~Zhou, M.~Redi, A.~Haines, and M.~Lalmas, ``Predicting pre-click quality for
  native advertisements,'' in \emph{Proceedings of the 25th International
  Conference on World Wide Web}, 2016, pp. 299--310.

\bibitem{guyon2004result}
I.~Guyon, S.~Gunn, A.~Ben-Hur, and G.~Dror, ``Result analysis of the nips 2003
  feature selection challenge,'' \emph{Advances in neural information
  processing systems}, vol.~17, 2004.

\bibitem{platt1998sequential}
J.~Platt, ``Sequential minimal optimization: A fast algorithm for training
  support vector machines,'' 1998.

\bibitem{cheng2016wide}
H.-T. Cheng, L.~Koc, J.~Harmsen, T.~Shaked, T.~Chandra, H.~Aradhye,
  G.~Anderson, G.~Corrado, W.~Chai, M.~Ispir \emph{et~al.}, ``Wide \& deep
  learning for recommender systems,'' in \emph{Proceedings of the 1st workshop
  on deep learning for recommender systems}, 2016, pp. 7--10.

\bibitem{wang2017deep}
R.~Wang, B.~Fu, G.~Fu, and M.~Wang, ``Deep \& cross network for ad click
  predictions,'' in \emph{Proceedings of the ADKDD'17}, 2017, pp. 1--7.

\bibitem{wang2021dcn}
R.~Wang, R.~Shivanna, D.~Cheng, S.~Jain, D.~Lin, L.~Hong, and E.~Chi, ``Dcn v2:
  Improved deep \& cross network and practical lessons for web-scale learning
  to rank systems,'' in \emph{Proceedings of the Web Conference 2021}, 2021,
  pp. 1785--1797.

\end{thebibliography}
}

\newpage
\appendix

\section{Details on Meta Optimization}
    We present the \textit{MetaAug} Algorithm~\ref{alg:maml} that operates on the unfolded concepts.
    It first computes the augmented concept vocabulary by taking every element in the global concept vocabulary except corresponding causal masks (line 2),
    by precomputing meta-data-set using concept alignment $\mathscr{D}_{Aug}$ from individual tasks (line 3-15), the optimization can be performed by iteratively sample mini-batches and back-propagate into meta-parameters using first order gradient (line 16 - 19) until convergence.
    After the meta-optimization stage for obtaining the meta-parameter $\omega^*$, we can either use the shared architecture \autoref{eq:theta} for deployment or perform fine tune on the specific dataset.

    \begin{algorithm}[H]
    \caption{\textit{MetaAug} Algorithm}
    \label{alg:maml}
    \begin{algorithmic}[1]
    \REQUIRE 
dataset collection of every tasks $\mathscr{D}  \defeq \{ (\mathcal{D}^{train~(i)}, \mathcal{D}^{val~(i)},  \mathcal{D}^{test~(i)}\}_{i=1}^K \} $ with global concept vocabulary $\mathcal{C}^{meta} $,
causal mask for each task $\mathbf{CMask}(\cdot)$
    \REQUIRE $\eta$: step size hyperparameters 
\STATE $\mathscr{D}_{Aug} \leftarrow $ empty set 
    \STATE $\mathcal{C}_{Aug}^i \defeq \mathcal{C}^{meta} \setminus  \mathbf{CMask}(i)$
    \FORALL{$\task_i$}
        \STATE update $\mathcal{C}_{Aug}^i$  according to \autoref{eq:caug}
        \FORALL{$split$ in $\{train, val, test\}$}
        \STATE $\mathcal{D}^{split~(i)}_{Aug} \leftarrow $ empty set
            \FORALL{$s \in \mathcal{D}^{split~(i)}_{Aug}$}
                \FORALL{$x\in \mathcal{C}_{Aug}^i$}
                \STATE $(\vec{c}_{s~Aug})_{x} \leftarrow $ association between entity $c$ and unfolded concept $x$ 
                \ENDFOR
            \STATE $(\mathcal{D}^{split~(i)}_{Aug}  \leftarrow \mathcal{D}^{split~(i)}_{Aug} \bigcup \{ (\vec{c}_{s~Aug}) \}$ 
            \ENDFOR
            \STATE  $\mathscr{D}_{Aug} \leftarrow \mathscr{D}_{Aug} \bigcup \mathcal{D}^{split~(i)}_{Aug}$
        \ENDFOR
        
    \ENDFOR
    \WHILE{not done}
    \STATE Sample batch of instance      $\mathcal{D}^{batch}$ from $ \bigcup_{ \mathcal{D} \in \{ 
     \mathcal{D}^{train~(i)} \in \mathscr{D}
     \} } \mathcal{D}$
     \STATE Update $\omega$ according to \autoref{eq:update}
    \ENDWHILE

    \end{algorithmic}
    \end{algorithm}

\section{Details on Online evaluation}
\begin{figure}[H]
\centering
\includegraphics[width=.5\textwidth]{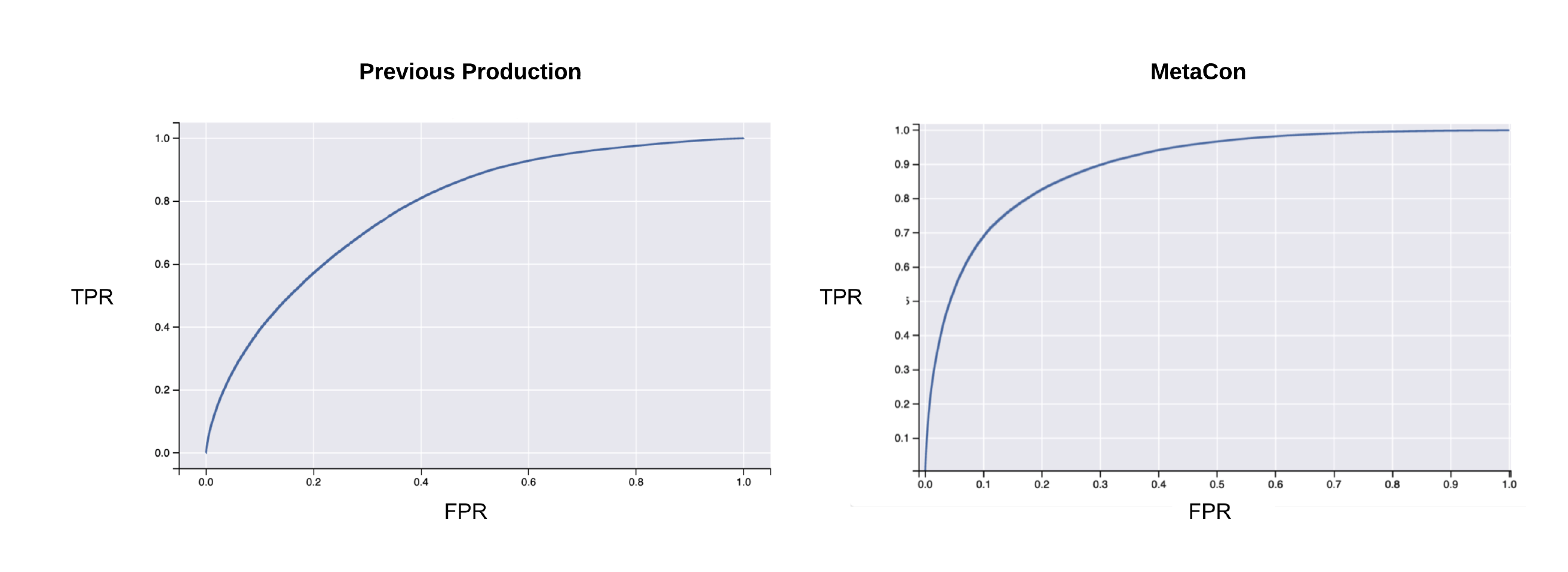}
\caption{Receiver operating curve of the online evaluation}
\label{fig-Online}
\vspace{10pt}
\end{figure}

The \textit{MetaCon} is deployed to online targeting use cases 
using Hadoop based deployment environment where we obtain performance  measurement on 
fresh online data. 
\autoref{fig-Online} shows the receiver operating curve of the online evaluation in key AUC-based targeting task between our system and the existing production system,
where we plot the true positive rate against false positive rate over all possible thresholds.
From the result, we can observer a consistent  improvement of our approach compared to production.

\end{document}